%% file: arxiv_main.tex
\documentclass[a4paper,10pt,pdftex]{article}
\usepackage{fullpage}
\usepackage{graphicx}
\usepackage[utf8]{inputenc}
\usepackage{amssymb,amsmath,amsthm,array,amsfonts}
\usepackage{color}
\usepackage{booktabs}
\usepackage{underoverlap}
\usepackage{subfigure}
\usepackage{multirow}
\usepackage{hyperref}
\usepackage[rightcaption]{sidecap}
\sidecaptionvpos{figure}{c}
\usepackage{tikz}
\usepackage[square,sort,comma,numbers]{natbib}
\usepackage[title]{appendix}
\usepackage{authblk}

\definecolor{matBlue}{rgb}{0, 0.4470, 0.7410}
\definecolor{matRed}{rgb}{0.8500, 0.3250, 0.0980}
\definecolor{matGreen}{rgb}{0.282, 0.635, 0.247}
\definecolor{matPurp}{rgb}{0.4940, 0.1840, 0.5560}

\newcommand{\tabitem}{~~\llap{\textbullet}~~}

\newcommand{\histplot}[1]{
    \begin{tikzpicture}[xscale=.5, yscale=.5, yshift=10cm]
        \begin{scope}[ycomb, yscale=1]
            \draw[black!80, line width=.5cm] plot #1;
        \end{scope}
    \end{tikzpicture}
}

\input{plot_data.txt}

\title{Large-scale detection and categorization of oil spills from SAR images with deep learning}

\author[1]{Filippo Maria Bianchi\thanks{filippombianchi@gmail.com}}
\author[2]{Martine M. Espeseth\thanks{martine.espeseth@uit.no}}
\author[1]{Njål Borch}

\affil[1]{\small NORCE the Norwegian Research Center}
\affil[2]{\small UiT the Arctic University of Norway}
\date{}

\begin{document}

\maketitle

\begin{abstract}
We propose a deep learning framework to detect and categorize oil spills in synthetic aperture radar (SAR) images at a large scale.
Through a carefully designed neural network model for image segmentation trained on an extensive dataset, we obtain state-of-the-art performance in oil spill detection, achieving results that are comparable to results produced by human operators.
We also introduce a classification task, which is novel in the context of oil spill detection in SAR. Specifically, after being detected, each oil spill is also classified according to different categories of its shape and texture characteristics.
The classification results provide valuable insights for improving the design of oil spill services by world-leading providers.
As the last contribution, we present our operational pipeline and a visualization tool for large-scale data, which allows to detect and analyze the historical presence of oil spills worldwide.
\end{abstract}

\section{Introduction}

Spaceborne Synthetic Aperture Radar (SAR) instruments have been used for monitoring and early detection of oil spills for several decades and are a well-established tool for many operational monitoring services. 
This information is paramount for planning oil spill preparedness strategies where location, extent, and early warning are relevant. 
Oil detections can concern both legal, illegal and accidental discharge from off-shore installations, ships, and pipelines. 
For example, in the North sea, oil slicks resulted from discharge of produced water\footnote{Produced water contains a small amount of mineral oil that consists of various constituents depending on the age and the type of oil well~\cite{Bakke2011,NIVA2019}.} are frequently detected around oil platforms~\cite{Skrunes2019,Bakke2011,NIVA2019}. 
The large amount of archived data of several thousands of detected and some verified oil spills over the years is a direct result of the long-term use of SAR. 
This has created a huge potential for new technologies and methods to emerge, which can take advantage of this work-effort and archived data. 
Remarkably relevant is the free database of satellite images obtained from the Sentinel-1 sensors, which is acquiring images since April 2014 (Sentinel-1A) and April 2016 (Sentinel-1B) \cite{sentinel1}. 
SAR sensors allow monitoring the surface independently of the weather and sun conditions. 
This is especially important in the North Sea and the Barents Sea due to heavy cloud cover and darkness for long periods of the year. 

In SAR images, the oil slicks appear as dark patches due to the low backscatter response compared to the surrounding clean sea areas. The low backscatter is a result of the oil damping of short-gravity and capillary ocean surface waves. 
The dark signature of oil slicks in SAR is also common for many other ocean features like low-wind areas and natural biogenic slick, also known as \textit{look-alikes}. 
An extensive effort has been made to design methodologies to distinguish oil slicks from natural biogenic slicks and/or low wind areas (see e.g., \cite{Skrunes2014,Salberg2019,Singha2013,Solberg1999,Krestenitis2019b,TOPOUZELIS2007264}). 
The backscatter of oil slicks depends on several factors like wind, sensor properties, and oil characteristics. 
For example, the wind is the main factor generating the ocean surface roughness and, therefore, wind significantly affect the oil-sea contrast. 
The oil-sea contrast also depends on the incidence angle of the satellite, as it affects the backscatter response.
In particular, both high and low incidence angles yield low oil-sea contrast \cite{ALPERS2017133,Gade1998}.
All these factors make it challenging to automatically detect, segment, and classify oil slicks. 

There are several works on oil slick segmentation and classification using SAR. 
Segmenting oil slicks or dark objects in SAR imagery has been performed for several decades, and most of the traditional oil slicks classification algorithms (see, e.g., \cite{Solberg1999,Frate2000,TOPOUZELIS2007264,Singha2013}) consist of three stages; (1) detection of dark formations; (2) feature selection; (3) statistical classification methods. 
A thorough review of the traditional and the early work regarding oil slick segmentation and classification methods was presented in 2005~\cite{BREKKE20051}. 
Early work by \cite{Frate2000} on distinguishing oil spills from look-alikes used a multilayer perceptron together with a visual inspection for areas of interest followed by features extraction. 
Topouzelis et al.~\cite{TOPOUZELIS2007264} used two neural networks for segmenting dark objects and then separating potential oil slicks from look-alikes, hence, avoiding visual inspection of the selected area. 
This framework has been adopted in several later studies~\cite{Singha2013,Pineda2013}. 

The application in remote sensing of deep learning models for computer vision outperformed the previous signal processing techniques and sets the new state-of-the-art in several tasks~\cite{zhu2017deep}. 
In the few last years, several works proposed to use convolutional neural networks (CNNs) to detect oil spills~\cite{Krestenitis2019a,Krestenitis2019b,CANTORNA2019105716}. 
Compared to traditional pattern recognition approaches, CNNs can be trained end-to-end, meaning that they learn from examples how to map input data into the desired output~\cite{goodfellow2016deep}. 
First, this greatly simplifies the task of the practitioner that is not required to design rules and specify critical hyperparameters (e.g., thresholds) to solve the inference task and that generalize well to unseen data.
Second, the practitioner is relieved from hand-crafting features that are, instead, automatically learned by a CNN by optimizing the training objective. 
Indeed, human-engineered features come with biases and are hampered by the limitations of humans in discovering complicated patterns and relationships in the data~\cite{zheng2018feature}. 
On the other hand, deep learning models are exceptionally data-hungry, especially if the architectures are large and have many trainable parameters.
To learn features and classification rules that are general enough and do not overfit the training data, it is necessary to expose the model to a large amount of input-output pairs, which are SAR images and segmentation masks in the case of oil spills detection.
Unfortunately, while unlabeled data are cheap and available in large quantity, labels are usually scarce and costly to obtain~\cite{bengio2012unsupervised}.
This is the reason why there are no examples of deep learning models trained on a large-scale dataset for oil spill detection.

The main objective of this paper is to develop a robust and automated framework to detect and classify oil slicks that will benefit operational monitoring services and oil spill preparedness authorities. 
Our contributions are summarized as follows.
\begin{itemize}
    \item (\textit{Detection}) we develop a CNN architecture that detects oil spills in SAR scenes with high accuracy. When trained on a large-scale dataset, our model achieves extremely high performance.
    \item (\textit{Classification}) each oil spill detected by our deep learning model is further processed by a second neural network, which classifies it according to shape, contrast, and texture categories.
    \item (\textit{Visualization}) we present our production pipeline to perform inference at a large scale and visualize the obtained results. Our visualization tool allows analyzing the presence of oil spills worldwide at specific times in history.
\end{itemize}

Closely related to our first contribution (detection), is the work presented in~\cite{Krestenitis2019b} that compares the performance of six existing CNN models for semantic segmentation in performing oil spill detection. 
Among the tested models, DeepLabv3+~\cite{chen2018encoder} achieves the best segmentation performance.
To train and evaluate the models, the authors introduce a new segmentation dataset based on the pollution events provided by the European Maritime Safety Agency (EMSA) through the CleanSeaNet service.
Compared to ours, the dataset is smaller as it consists of 1112 images, each one covering an area of approximately $(12.5 \times 6.5) km^2$.
Additionally, the images are associated with segmentation masks with 5 classes (sea, land, oil, ships, and look-alikes), while we use binary masks (oil and non-oil). 
Finally, differently from~\cite{Krestenitis2019b}, we do not adopt off-the-shelf architectures but rather propose a CNN model, a training, and evaluation procedure that are optimized for the oil spill segmentation task at hand.

The remainder of this paper is organized as follows. 
In Section~\ref{sec:data}, we describe the original data and the preparation of the dataset used for training the proposed deep learning framework. 
In Section~\ref{sec:detection}, we present the CNN architecture used to perform semantic segmentation of oil spills.
Section~\ref{sec:detection} describes the second CNN model that categorizes the oil spills after they are detected by the segmentation network.
In Section~\ref{sec:results} we report the performance of the proposed framework and discuss the results obtained in comparison to the golden standard of manual labelling and \textit{in-situ} measurements.
Section~\ref{sec:nlive} describes the software we developed for large-scale visualization and analysis, based on the proposed framework.
Finally, in Section~\ref{sec:conclusions} we draw our conclusions.

\section{Dataset description}
\label{sec:data}

The dataset has been produced by Kongsberg Satellite Service (KSAT)\footnote{\url{www.ksat.no}} and consists of Synthetic Aperture Radar (SAR) scenes from Sentinel-1, associated with a binary mask generated by trained human operators at KSAT, which indicates the location and extent of the oil spills.
KSAT has a long and consolidated experience in oil detection by SAR, offering worldwide near-real-time services with extensive coverage and fine temporal resolution based on different SAR sensors. 
The masks associate each pixel with a label that is 0 for the class ``non-oil spill'' and 1 for the class ``oil spill''. 
The whole dataset consists of 713 products of the Sentinel-1 sensor and each product covers an area up to approximately $150,000\; km^2$.
The products are collected over a period of 4 years between 2014 and 2018, and they contain $2,093$ \textit{oil spill events}. An oil spill event refers to a collection of one or several single oil slicks that are located nearby and originate from the same source according to trained human operators at KSAT.
The total number of individual oil spills is $227,964$. 

The SAR scenes are acquired with the dual-polarimetric SAR mode (IW mode of Sentinel-1) using the vertical transmit and vertical receive (VV) and vertical transmit and horizontal receive (VH) polarization channels with the medium resolution mode (see \cite{sentinel1} for additional information about Sentinel-1). 
The VV polarization is preferred over VH for oil spill detection, due to less impact of system noise in VV compared to the VH channel. 
For this reason, the VV channel is the one considered in this study. 

All the SAR products used for training the deep learning models are smoothed for noise removal and are at 40 meters resolution, \textit{i.e.}, each pixel covers an area of $40 \times 40$ meters.
After the model is trained, to perform detection also on the high-definition (10m resolution) SAR products that are publicly available from the Copernicus Open Access Hub (\url{https://scihub.copernicus.eu/}), we first applied a boxcar filter with size 11 and then we down-sampled the images to 1/4 of their original size to recover the 40m resolution.

The values in the VV channel are provided in 16bit unsigned integer format, meaning that the backscatter assumes values in the interval $[0, 2^{16}]$. 
Radiometric calibration to sigma-nought is not performed since is a linear transformation and we expect the deep learning model to learn to apply such a transformation if needed.
Very few outlier pixels (typically backscattering from ships/platforms or pixels located at high incidence angle) in the dataset have high backscatter values up to $2^{16}$, while the majority are concentrated around a much smaller interval. 

\begin{SCfigure}[.5][!ht]
    \centering
    \includegraphics[keepaspectratio,width=0.5\textwidth]{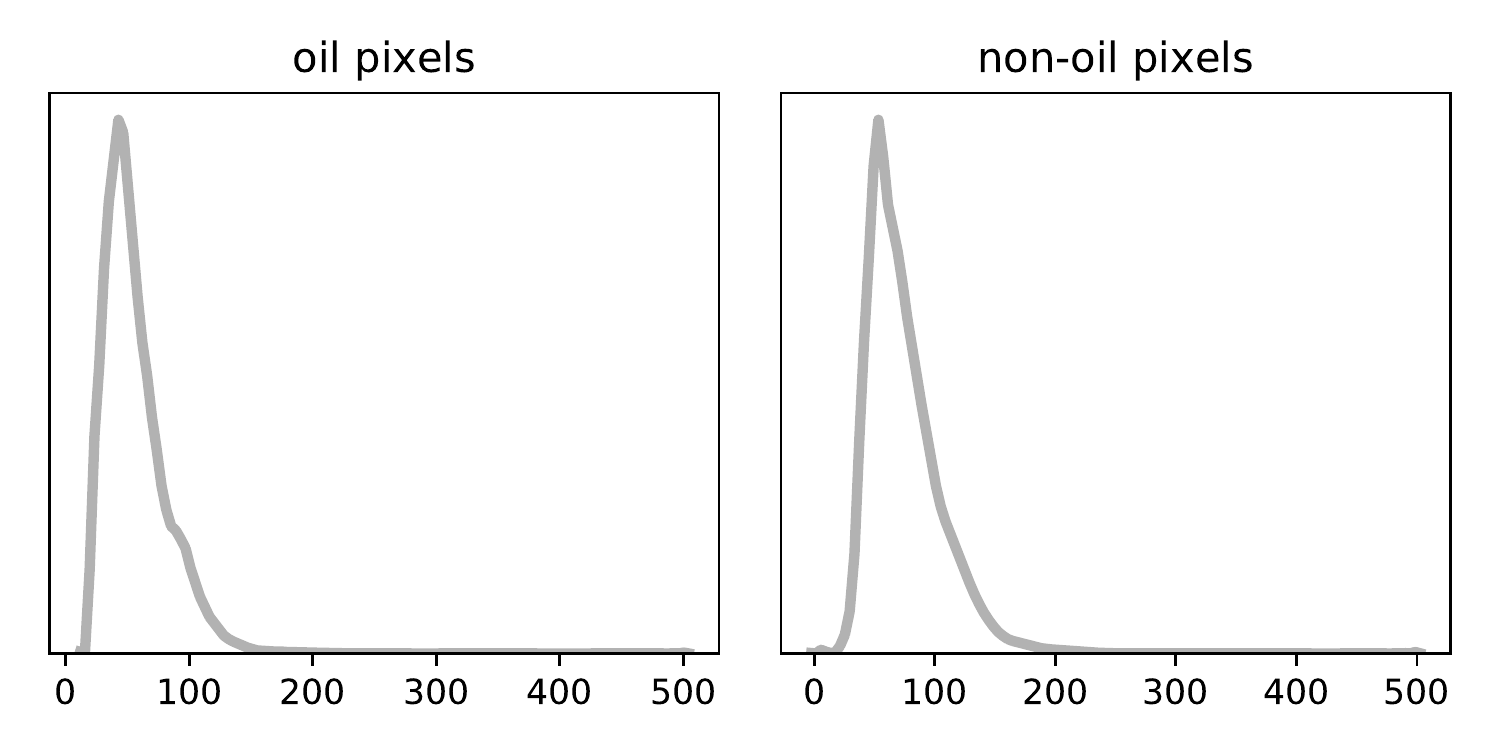}    
    \caption{Distribution of the backscatter values in the VV images, across pixels belonging to class ``oil'' and ``non-oil''. 
    The plots show only the range $[0,500]$, rather than $[0, 2^{16}]$.}
    \label{fig:distr}
\end{SCfigure}

Fig.~\ref{fig:distr} shows a normalized distribution of the values of the pixels marked as ``oil'' and ``non-oil'' computed over all the SAR products in our dataset. Since pixels with high backscatter are very few, for visualization purposes the density plots show a limited range $[0,500]$ of backscatter values, rather than the whole interval $[0, 2^{16}]$. 
As we can see from the plots, the backscatter distribution is shifted toward smaller values. 
As expected, the backscatter distribution of oil-covered pixels is shifted towards values that are lower than non-oil pixels. 
Based on this analysis, we clip the maximum value in all VV images in the dataset to 150, which corresponds to approximately the 98\% percentile.
Notably, all the oil spill pixels have a backscatter value lower than 150.

Besides the binary masks indicating the position and shape of oil spills, each oil spill event in the dataset is categorized according to 12 different fields, which indicate the type of shape and texture of the oil spill. 
Tab.~\ref{tab:oil_categories} reports the name of the 12 categories, the set of possible values assumed by each category, and the distribution of the values across the dataset.
It is possible to notice immediately that for several categories, such as the texture attributes, the distribution of the values is very skewed.
This likely indicates that discriminating across certain categories is challenging for the human operators that are labelling the SAR scenes.
In particular, texture categories are difficult to detect given the inherent noise in the VV channel.
An objective in this study is to assess if such categories can be predicted using machine learning; indirectly this indicates how much information about texture, shape, and contrast can be extrapolated from the SAR image.

\begin{table}
\caption{The 12 categories used to classify each oil spill. The possible values assumed by each category are reported in the second column. The distribution the values in each category is shown in the third column.}
\setlength\tabcolsep{.6em} 
\footnotesize
\bgroup
\def\arraystretch{.01} 
\centering
\begin{tabular}{l|c|cc}
\cmidrule[1.5pt]{1-4}
\textbf{Category} & \textbf{Values} & \multicolumn{2}{c}{\textbf{Values distribution}} \\
\cmidrule[.5pt]{1-4}
\multirow{1}{*}[4.5ex]{Patch shape} & \multirow{1}{*}[4.5ex]{False, True} & \multirow{1}{*}[4.5ex]{[0.482, 0.518]} & \histplot{coordinates {\patchDATA}} \\
\multirow{1}{*}[4.5ex]{Linear shape} & \multirow{1}{*}[4.5ex]{False, True} & \multirow{1}{*}[4.5ex]{[0.618, 0.382]} & \histplot{coordinates {\linearDATA}} \\
\multirow{1}{*}[4.5ex]{Angular shape} & \multirow{1}{*}[4.5ex]{False, True} & \multirow{1}{*}[4.5ex]{[0.904, 0.096]} & \histplot{coordinates {\angularDATA}}\\
\multirow{1}{*}[4.5ex]{Weathered texture} & \multirow{1}{*}[4.5ex]{False, True} & \multirow{1}{*}[4.5ex]{[0.713, 0.287]} & \histplot{coordinates {\weatheredDATA}} \\
\multirow{1}{*}[4.5ex]{Tailed shape} & \multirow{1}{*}[4.5ex]{False, True} & \multirow{1}{*}[4.5ex]{[0.832, 0.168]} & \histplot{coordinates {\tailedDATA}}\\
\multirow{1}{*}[4.5ex]{Droplets texture} & \multirow{1}{*}[4.5ex]{False, True} & \multirow{1}{*}[4.5ex]{[0.983, 0.017]} & \histplot{coordinates {\dropletsDATA}}\\
\multirow{1}{*}[4.5ex]{Winding texture} & \multirow{1}{*}[4.5ex]{False, True} & \multirow{1}{*}[4.5ex]{[0.925, 0.075]} & \histplot{coordinates {\windingDATA}}\\
\multirow{1}{*}[4.5ex]{Feathered texture} & \multirow{1}{*}[4.5ex]{False, True} & \multirow{1}{*}[4.5ex]{[0.977, 0.023]} & \histplot{coordinates {\featheredDATA}}\\
\multirow{1}{*}[4.5ex]{Shape outline}  & \multirow{1}{*}[4.5ex]{Fragmented, Continuous} & \multirow{1}{*}[4.5ex]{[0.784, 0.216]} & \histplot{coordinates {\shapeDATA}} \\
\multirow{1}{*}[4.5ex]{Texture} & \multirow{1}{*}[4.5ex]{Rough, Smooth, Strong, Variable} & \multirow{1}{*}[4.5ex]{[0.292, 0.143, 0.051, 0.512]} & \histplot{coordinates {\textureDATA}} \\
\multirow{1}{*}[4.5ex]{Contrast} & \multirow{1}{*}[4.5ex]{Strong, Weak, Variable} & \multirow{1}{*}[4.5ex]{[0.230, 0.528, 0.241]} & \histplot{coordinates {\contrastDATA}}\\
\multirow{1}{*}[4.5ex]{Edge} & \multirow{1}{*}[4.5ex]{Sharp, Diffuse, Variable}  & \multirow{1}{*}[4.5ex]{[0.337, 0.133, 0.529]} & \histplot{coordinates {\edgeDATA}}\\
\cmidrule[1.5pt]{1-4}
\end{tabular}
\label{tab:oil_categories}
\egroup
\end{table}

\subsection{Division in patches}
\begin{figure}[!ht]
    \centering
    \includegraphics[keepaspectratio,width=\textwidth]{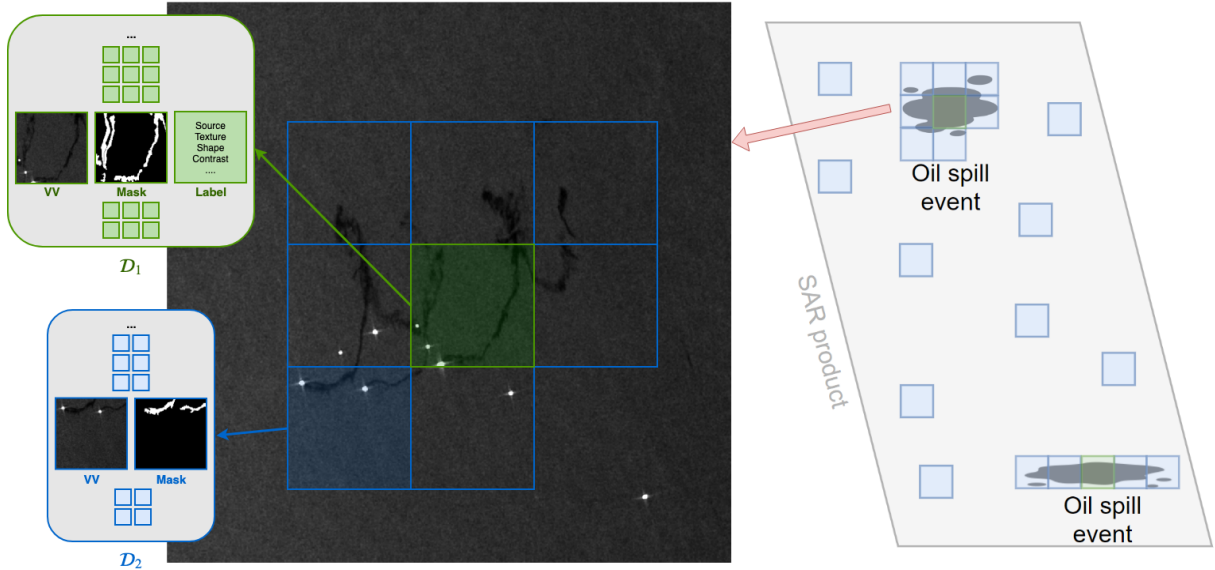}    
    \caption{Illustration of patches extraction from the SAR products.
    The patches centred on the oil spill events (depicted in green) form the dataset $\mathcal{D}_1$ and are associate with a label that indicates the values of the 12 categories for that oil event.
    The second dataset, $\mathcal{D}_2$, contains i) all the patches of $\mathcal{D}_1$, all the patches with at least 1 oil spill pixel, iii) an equal amount of patches without oil, randomly sampled from other locations in the SAR product.
    Along with the VV channel, the segmentation masks are always included in both datasets.}
    \label{fig:tiling}
\end{figure}

To convert the dataset in a format suitable for training a CNN architecture, we extracted a set of patches of size $160 \times 160$ pixels from the SAR products, each one covering an area of $41\; km^2$.
An entire oil spill event can be very large and a patch, in general, does not cover it completely. Recall, an oil spill event can cover several single neighbouring oil slicks originating from the same source.
By referring to the example in Fig.~\ref{fig:tiling}, 8 patches are necessary to cover the oil event depicted in the figure.
Since the labels in Tab.~\ref{tab:oil_categories} are associated with a whole oil event, which can be composed of multiple oil slicks, two options can be considered to associate labels and patches. 
The first is to assign the same label to all the patches covering the same oil event.
The second is to take the most central of the patches covering the oil event (depicted as the green box in Fig.~\ref{fig:tiling}) and the label of the whole oil event is only assigned to the central patch.
We opted for this second option and ended up with a total of $2,093$ ``centred'' patches associates with a label describing the values of the 12 attributes in Tab.~\ref{tab:oil_categories}.
We refer to this dataset as $\mathcal{D}_1$.

We note that $\mathcal{D}_1$ includes only a fraction of the available segmentation masks: by referring to the oil event depicted in Fig.~\ref{fig:tiling}, all the data outside the green central patch is not contained in $\mathcal{D}_1$.
To exploit also the remaining information (i.e., the segmentation masks associated to the oil spill pixels outside the green box in Fig.~\ref{fig:tiling}), we built a second dataset $\mathcal{D}_2$, which includes all the patches of $\mathcal{D}_1$ plus all the patches that contain at least one pixel belonging to the oil class.
The additional patches assigned to $\mathcal{D}_2$ are depicted as blue boxes in Fig.~\ref{fig:tiling}.
We note that all the patches in $\mathcal{D}_1$ (green boxes) are also included in $\mathcal{D}_2$.
During the training phase, we want to expose the segmentation model also to patches where no oil is present. 
Therefore, we also included in $\mathcal{D}_2$ patches without any oil spill pixels, which are randomly sampled from the SAR products.
The total number of patches in $\mathcal{D}_2$ is $187,321$ and approximately half of the patches do not contain any oil pixel.
The total amounts of pixels belonging to class 0 and 1 are $59,689,609$ and $1,243,242$, respectively. 
Therefore, the pixels of class ``oil'' is $2.04\%$ of the total.

\subsection{Division on in training, validation, and test set.}
The dataset $\mathcal{D}_1$ is used to train the deep learning model that performs classification (see Sect.~\ref{sec:classification}) and to perform hyperparameter selections (see Sect.~\ref{sec:results}).
We split $\mathcal{D}_1$, in training, validation, and test set with sizes $1843$, $150$, and $100$ respectively.

The dataset $\mathcal{D}_2$ is used to train the model that performs oil spills detection and is split in a training and validation set of sizes $149,856$ and $37,465$, respectively.
We made sure that all the patches in the validation and test set of $\mathcal{D}_1$ are excluded from the training set of $\mathcal{D}_2$.

From the original dataset consisting of $713$ SAR products, 3 whole products are kept aside, \textit{i.e.} they are not used to extract the patches that populate $\mathcal{D}_1$ or $\mathcal{D}_2$.
These 3 products form a separate dataset, $\mathcal{D}_\text{test}$, which is used exclusively to test the performance of the segmentation task.
Tab.~\ref{tab:dataset_details} summarizes the content of the datasets $\mathcal{D}_1$, $\mathcal{D}_2$, and $\mathcal{D}_1$ or $\mathcal{D}_\text{test}$.
Sect.~\ref{sec:results} has a special focus on the performance achieved on the three test scenes. Tab.~\ref{tab:test_data} provides additional details on $\mathcal{D}_\text{test}$ that will be useful for discussing the results.

\bgroup
\def\arraystretch{1} 
\begin{table}
  \centering
  \small
  \begin{tabular}{llll}
  \cmidrule[1.5pt]{1-4}
   \textbf{ Original data} & $\mathcal{D}_1$ & $\mathcal{D}_2$ & $\mathcal{D}_\text{test}$ \\
    \midrule
    \tabitem 713 SAR prod. & \tabitem 1,843 tr. patches & \tabitem 149,856 tr. patches & \tabitem 3 SAR products \\
    \tabitem 4 years period     & \tabitem 150 val. patches  & \tabitem 37,465  val. patches & \tabitem Details in Tab.~\ref{tab:test_data}\\
    \tabitem 2,093 oil events   & \tabitem 100 test patches & & \\
    \tabitem 227,964 oil spills &  & & \\
  \cmidrule[1.5pt]{1-4}
  \end{tabular}
  \caption{Summary of the datasets details.}
  \label{tab:dataset_details}
\end{table}
\egroup

\bgroup
\def\arraystretch{1} 
\setlength\tabcolsep{1.0em} 
\begin{table}[!ht]
\small
\centering
\begin{tabular}{lcccc}
\cmidrule[1.5pt]{1-5}
\textbf{ID} & \textbf{Pixel size} & \textbf{\# oil spills} & \textbf{\# oil pixels} & \textbf{\# non-oil pixels}\\
\cmidrule[.5pt]{1-5}
T1 & $9,836 \times 14,894$ & 2 & 552 (0.00067\%) & 81,877,550 \\ 
T2 & $9,470 \times 21,738$ & 11 & 19,336 (0.017\%) & 112,071,385 \\ 
T3 & $10,602 \times 21,471$ & 36 & 22,793 (0.018\%) & 127,436,689 \\ 
\cmidrule[1.5pt]{1-5}
\end{tabular}
\caption{Further details on $\mathcal{D}_\text{test}$, \textit{i.e.}, the three SAR products used as test set.}
\label{tab:test_data}
\end{table}
\egroup


\section{Oil spill detection}
\label{sec:detection}
The oil spill detection is conveniently framed as a semantic segmentation task, which consists in performing pixel-level classification~\cite{zhao2017survey}.
In the following, we first describe the neural network model used to perform segmentation, then the procedures adopted for training the model and to perform inference on new, unseen data. 

\subsection{The deep learning architecture for semantic segmentation}
\label{sec:architecture}
The model used to perform segmentation is a fully convolutional network, referred in the rest of the paper as \textit{OFCN} (Oil Fully ConvNet).
The OFCN is a network with no dense layers, which can process inputs of variable size.
This allows for training on small images and processing larger ones at inference time.
The OFCN model is based on the U-net~\cite{ronneberger2015u}, a popular deep learning architecture for image segmentation that is also used in several remote sensing applications~\cite{ghosh2018stacked, li2018deepunet, bianchi2019snow}.

\begin{figure}[!ht]
    \centering
    \includegraphics[keepaspectratio,width=0.7\textwidth]{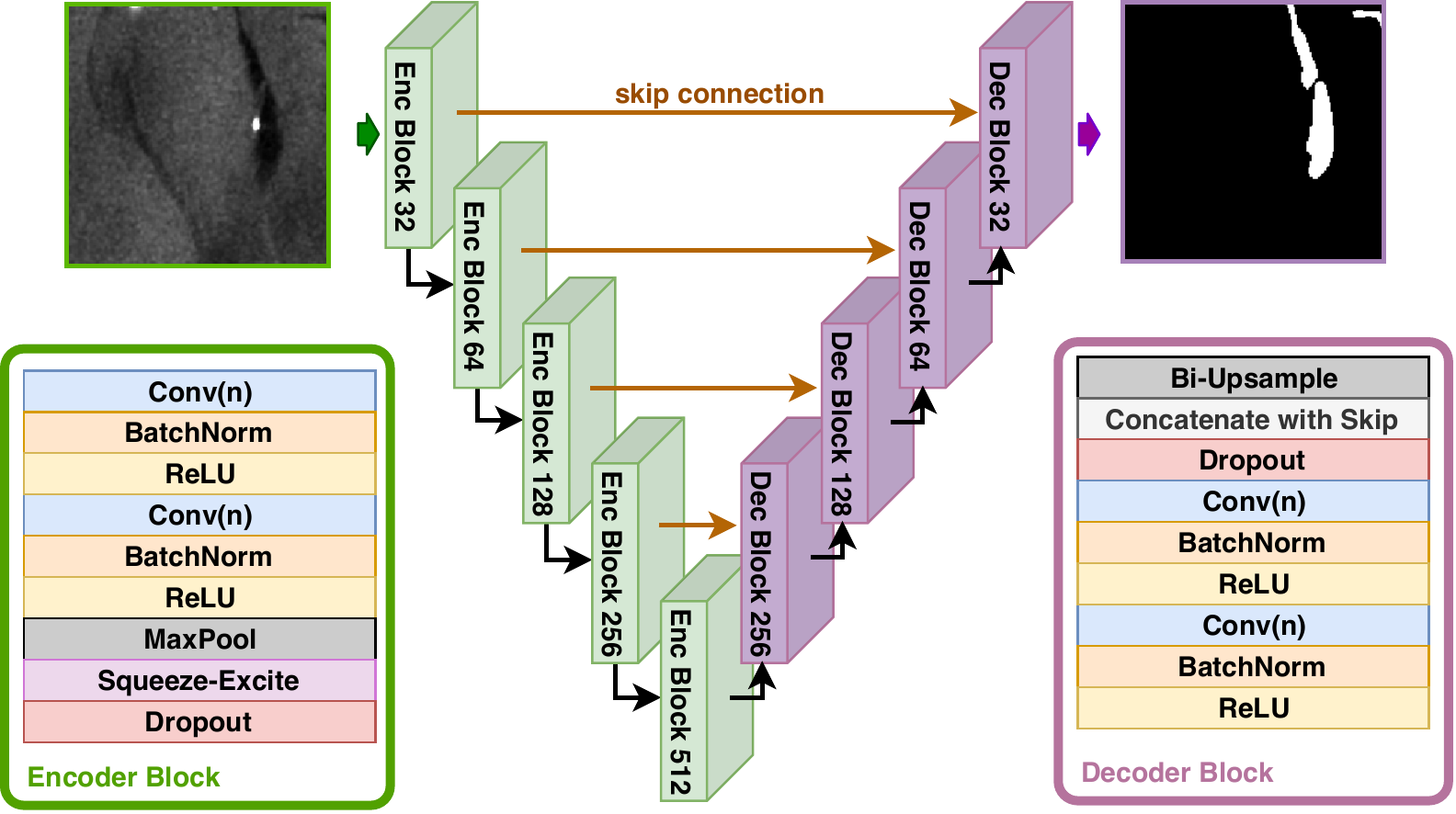}    
    \caption{Schematic depiction of the OFCN architecture used for segmentation. Conv($n$) stands for a convolutional layer with $n$ neurons. For example, $n=32$ in the first Encoder Block, $64$ in the second, and so on. }
    \label{fig:unet}
\end{figure}
The OFCN consists of an \textit{encoder} and a \textit{decoder} part, respectively depicted in green and purple in Fig.~\ref{fig:unet}.
The encoder gradually extracts feature maps that detect the patterns of interest in the image.
By reducing the spatial dimensions and increasing the number of filters, the deeper layers in the encoder capture features of increasing complexity and larger spatial extent in the input image.
The decoder gradually transforms the high-level features and, in the end, maps them into the output.
The output is a binary segmentation mask, which has the same height/width of the input image and associates to each pixel a class value: 1 if it belongs to the oil class, 0 otherwise.
The skip connections link the feature maps from the encoding to the decoding layers, such that some information can bypass the bottleneck located at the bottom of the architecture.
In this way, our architecture still learns to generalize from the high-level latent representation but also recovers spatial information from the intermediate representations through a pixel-wise semantic alignment.

Each block in the encoder consists of 9 layers, as depicted in the green box of Fig.~\ref{fig:unet} (bottom-left).
Conv($n$) indicates a convolutional layer with $n$ filters (\textit{e.g.}, $n = 32$ in the 1st block, $64$, $128$, $256$, and $512$ in the 2nd, 3rd, 4th and 5th blocks, respectively), filter size $3 \times 3$, stride $1$, and padding modality \textit{same}~\cite{dumoulin2016guide}.
Each Conv layer is followed by a Batch Normalization (BN) layer~\cite{ioffe2015batch} and a ReLU activation function.
At the end of each block, there is a max-pooling with stride $2$, a Squeeze-and-Excitation (SE)~\cite{hu2018squeeze}, and a Dropout layer~\cite{srivastava2014dropout}.
Each unit in the deeper layer of the encoder has a receptive field of 140, meaning that each feature depends on a neighbourhood with a radius of 140 pixels in the input image.
A visualization of the growth of the receptive field in the encoder layers is reported in Appendix~\ref{sec:model_details}.

Compared to the encoder, the decoder has a somehow mirrored structure and the details of each block are shown in Fig.~\ref{fig:unet} (bottom-right). 
The main differences are the Bilinear upsampling layers, which upscale the features map of the previous layer, and concatenation with the skip connections that injects in the decoder the output of the encoder blocks.
The last decoder block replaces the second ReLU activation with a sigmoid that produces output in the interval $[0, 1]$.
This is a common choice in binary classification tasks, such as the generation of the binary oil spill mask.

Let $n$ be the number of convolutional filters in the first layer, the number of filters in the rest of the OFCN architecture is univocally determined and is $n$-$(n\times2)$-$(n\times4)$-$(n\times8)$-$(n\times16)$-$(n\times8)$-$(n\times4)$-$(n\times2)$-$n$. 
For conciseness, the notation OFCN($n$) in the rest of the paper is used,
For example, the architecture in Fig.~\ref{fig:unet} is OFCN(32).

In the Appendix, we describe Bilinear upsampling, Batch Normalization, Squeeze-and-Excitation, and Dropout modules.

\subsection{Training and inference}
\label{sec:training}
The OFCN is trained to predict the binary segmentation mask, indicating the presence of oil spills.
The network weights are optimized by iteratively minimizing a loss function evaluated on mini-batches of input-output pairs.
The choice of the loss function and the other training procedures are reported in the following.

\subsubsection{Loss functions for unbalanced dataset}
Oil spills are small objects and, as discussed in Sect.~\ref{sec:data}, they represent only a tiny fraction ($\approx 2\%$) of the entire dataset.
Due to the strong imbalance between the pixels of class 0 (``non-oil spill'') and class 1 (``oil spill''), a naive classifier can achieve an accuracy of $\approx 98\%$ simply by assigning all the pixels to class 0.
To handle the class imbalance, rather than training the OFCN with the standard binary cross-entropy loss, a binary cross-entropy with class balancing is used.
Specifically, the standard cross-entropy is re-weighted to assign a larger penalty when a pixel of the under-represented class is wrongly classified. 
This can be done by weighting the loss associated with each pixel of the oil class with a value larger than for the non-oil class.
We also tested three additional loss functions that account for class imbalance but obtained unsatisfactory results.
The details are in Appendix~\ref{sec:not_working}.

\subsubsection{Data augmentation} 
To prevent the model from overfitting the training data and to enhance its generalization capability on unseen examples, we augment the dataset during training.
Randomized data augmentation can improve the generalization performance in several computer vision tasks, including applications on remote sensing~\cite{ding2016convolutional}.
In particular, we apply the following random transformations on the fly: horizontal and vertical flips, horizontal and vertical shifts, rotations, zooming and shearing to the training images. 
To ensure consistency between input and the target segmentation masks used for training, the same transformations of the input are also applied to the oil masks.

\subsubsection{Two-stage training}
We first trained the OFCN on a low-resolution version of the dataset, obtained by downsizing the patches size by half ($80 \times 80$ pixels).
After having completed the training on the low-resolution dataset, we resumed the training on the patches of original size \textit{without resetting the network weights}.
This is possible because, as discussed in Sect.~\ref{sec:architecture}, the OFCN can consume images of variable size, since its weights are independent of the input shape.
The intuition behind training first on downsized images is to let the model learn first the coarser structure in the inputs and then refine the parameters' tuning as the incoming images expand and become more detailed.

\subsubsection{Test time augmentation.}
When computing the prediction of a SAR scene at inference time, we \textit{slide} the OFCN on the large image, computing predictions for one window at a time.
Again, we stress that the window size at test time can be larger than $160 \times 160$ pixels.
However, this approach usually generates checkerboard artefacts and border effects close to the window edges.
To obtain smoother and more accurate predictions, overlapping sliding windows with a stride equal to half the window size are processed by the OFCN.
Furthermore, 8 predictions from all the possible 90$^{\circ}$ rotations and flips of each window are generated.
To obtain the final result, we used a $2^\text{nd}$ order spline interpolation to merge all the computed predictions.

\section{Oil spill classification}
\label{sec:classification}
\begin{figure}[!ht]
    \centering
    \includegraphics[keepaspectratio,width=.75\textwidth]{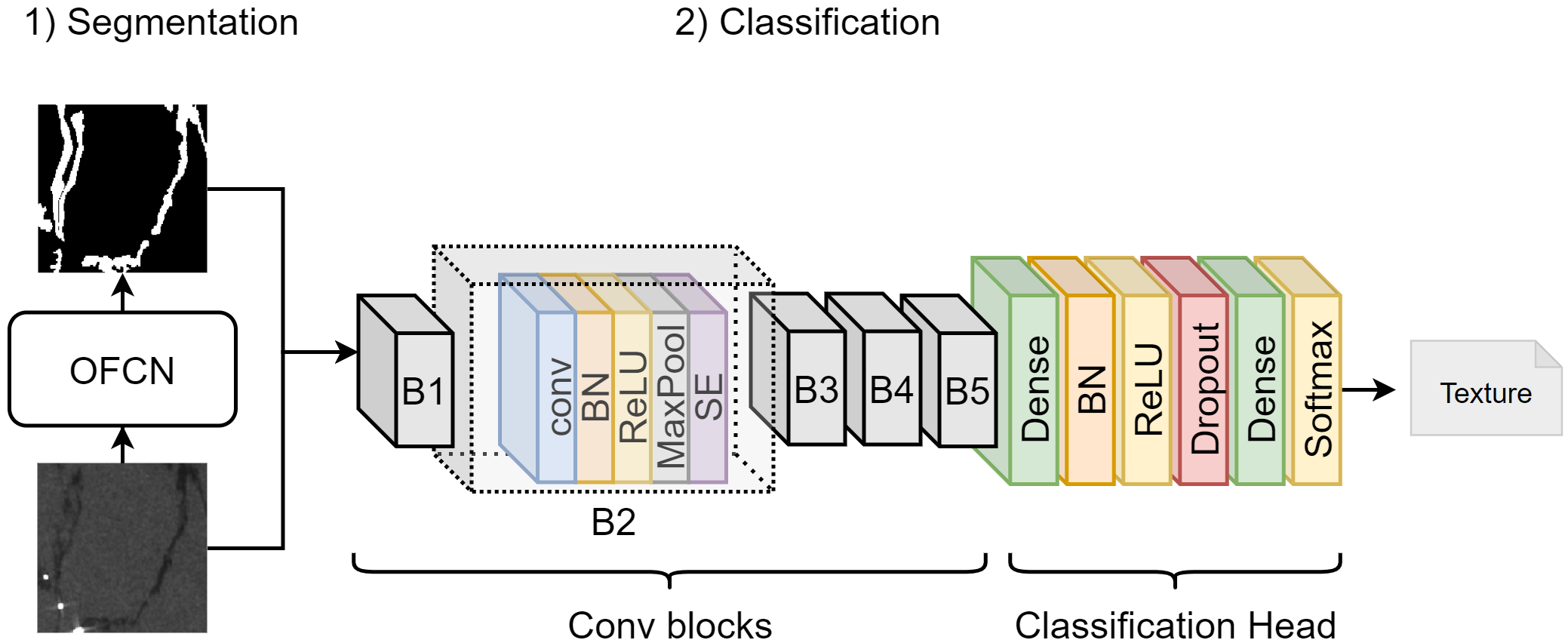}    
    \caption{First the trained OFCN generates the segmentation masks fro the SAR images. 
    Then, both SAR images and predicted mask are fed in the classification network. A different architecture is trained to classify each one of the categories (e.g., the depicted one classifies the ``texture'' category).}
    \label{fig:oilclass}
\end{figure}

With \textit{oil spill classification} we refer to the task of predicting the 12 categories, described in Tab.~\ref{tab:oil_categories}, which indicate the texture, shape, and contrast of the oil spill. 
Such categories are useful to end-users and analysts in deriving information about the source, stage of weathering and internal variations within the oil slicks. 
To classify each of the 12 categories, we train a separate instance of the architecture depicted in Fig.~\ref{fig:oilclass}.
The model takes as input a SAR patch and the associated mask predicted by the OFCN model described in Sect.~\ref{sec:detection}.
The (predicted) segmentation mask encourages the classification network to focus on the areas of the patch with oil spills and allows to extract more easily shape information.

Fig.~\ref{fig:oilclass} shows the whole pipeline: input SAR image $\rightarrow$ detection $\rightarrow$ classification.
The pipeline is not trained end-to-end. In fact, we first train the OFCN and, afterwards, the classification network.
The first, obvious reason for not training the whole pipeline end-to-end is that the category labels are available only for $\mathcal{D}_1$ and not $\mathcal{D}_2$, the large dataset used to train the OFCN. 
Moreover, compared to the segmentation masks the 12 category labels are noisier as they are more subjective to human interpretation. 
Therefore, to achieve the best possible segmentation performance, the OFCN is trained independently without being conditioned by the classification loss.

The classification network in Fig.~\ref{fig:oilclass} consists of 5 convolutional blocks responsible for feature extraction.
Each block is structured as [conv($n$)-BN-ReLU-Maxpool-SE], where the number of convolutional filters $n$ is 32, 64, 128, 256, and 512 in the blocks B1-B5, respectively.
The classification head has the following architecture: Dense(256)-BN-ReLU-Dropout-Dense(\#cat)-Softmax.
The layers ``Dense'' are two fully-connected layers whose numbers of units are 256 and the number of values assumed by each category (\#cat), respectively.
The network is trained by minimizing a categorical cross-entropy loss and by using the same image augmentation procedure used for training the OFCN architecture.

\section{Results and discussion}
\label{sec:results}

In this section, we describe the experimental setting and report the results obtained for the detection and classification task, respectively.

\subsection{Oil spill detection: experimental setting, analysis, and results}
\label{sec:detect_results}

\subsubsection{Evaluation metrics.}
While the parameters of OFCN are optimized by minimizing the loss described in Sec.~\ref{sec:detection}, we require more interpretable metrics to quantify the results obtained on the test, and also to monitor the model performance on the validation set during training.
We considered different metrics rather than accuracy since the oil class is highly under-represented in the segmentation task and also some categories are imbalanced in the categorization task.

The first metric is the F1 score, which is computed at the pixel level and is defined as
\[ F1 = 2 \frac{ \text{precision} \cdot \text{recall} }{ \text{precision} + \text{recall} },\]
where \textit{precision} is defined as $\frac{TP}{TP+FP}$ and \textit{recall} is $\frac{TP}{TP+FN}$ (TP = True Positives, FP = False Negatives, FN = False Negatives).
To compute the F1 score in the segmentation task, the output $o$ of the sigmoid in the last layer of the OFCN is rounded as follows: $o = 1$ if $o \geq \tau$, $o = 0$ if $o < \tau$.
When not specified otherwise, we use $\tau=0.5$.

In each experiment presented in the following, the F1 score on the validation set is used to evaluate during training the performance of the current model on unseen data. 
Specifically, whenever the model improves its F1 score on validation, the current instance of the weights are saved as the best model.

We also consider a second metric that indicates if the OFCN managed to correctly locate the oil spill, without accounting for small differences in the shape contours of human-made and predicted segmentation masks.
For this purpose, we consider the bounding boxes that contain oil spills in both the human-made and predicted mask.
TPs are now measured as the number of bounding boxes in the human-made mask that has a non-zero intersection with a bounding box in the predicted mask.
Similarly, we compute the FP and FN.
To quantify how much the bounding boxes in the ground truth and the prediction overlap, we computed the intersection over union (IoU):
\[IoU = \frac{\text{Area of bounding boxes intersection}}{\text{Area of bounding boxes union}}.\]

Contrarily to the F1 score that is evaluated during training to save the best model, the IoU is only computed once the training is over to evaluate the final performance. 

\subsubsection{Hyperparameters search}
To find the optimal configuration of the hyperparameters in the OFCN, we performed cross-validation by randomly sampling configurations from the hyperparameters space and selecting those that yield the highest performance on the validation set.
A total of $500$ configurations are sampled from the following hyperparameters space: BN \{True, False\}, SE \{True, False\}, Loss \{weighted binary cross-entropy, Jaccard, Focal, Lov{\'a}sz-softmax\}, L\textsubscript{2} penalty \{0, 1e-6, 1e-5, 1e-4, 1e-3, 1e-2\}, Dropout \{0, 0.1, 0.25, 0.5\}, learning rate of Adam~\cite{kingma2014adam} optimizer \{1e-4, 5e-4, 1e-3, 5e-3\}, oil class weight (only binary cross-entropy) \{1, 2, 3, 5\}\footnote{We also tried weights higher than 5 for the oil class, but obtained a very large number of false positives.}.

To make the hyperparameters search tractable, we used a smaller architecture, OFCN(16), and we trained it on the training and validation set of $\mathcal{D}_1$, which is much smaller than $\mathcal{D}_2$, for 100 epochs only.
We used mini-batches of size 32 and image augmentation is used with the following parameters: 
max rotation 90$^{\circ}$, 
max width shift 0.1 of total width, 
max height shift 0.1 of total height, 
max shearing 0.3, 
max zoom 0.2, 
probability of horizontal and vertical flips 0.5, 
pad mode ``mirror''.
Tab.~\ref{tab:crossval} reports the F1 score obtained by the 3 best configurations. 

\bgroup
\def\arraystretch{1} 
\setlength\tabcolsep{1em} 
\begin{table}[!ht]
\footnotesize
\centering
\begin{tabular}{lccccccc}
\cmidrule[1.5pt]{1-8}
\textbf{ID} & \textbf{BN} & \textbf{SE} & \textbf{L\textsubscript{2} reg.} & \textbf{Dropout} & \textbf{LR} & \textbf{CW} & \textbf{Var F1 ($\mathcal{D}_1$)} \\
\cmidrule[.5pt]{1-8}
\textbf{C1} & True & True  & 0.0  & 0.1 & 1e-3 & 2 & \textbf{0.731} \\ 
\textbf{C2} & False & True & 1e-6 & 0.1 & 5e-4 & 3 & 0.723 \\
\textbf{C3} & True & False & 0.0  & 0.0 & 1e-3 & 2 & 0.708 \\
\cmidrule[1.5pt]{1-8}
\end{tabular}
\caption{Hyperparameters selection results. We report the 3 best configurations (C1, C2, C3) found with cross-validation on $\mathcal{D}_1$. \textit{Acronyms:} BN (Batch Normalization), SE (Squeeze-and-Excitation), L\textsubscript{2} reg. (strength of the L\textsubscript{2} regularization on the network parameters), LR (Learning Rate), CW (weight of the oil class). }
\label{tab:crossval}
\end{table}
\egroup

\subsubsection{Comparison with baselines}
We compare the performance of the proposed OFCN with the vanilla U-net architecture~\cite{ronneberger2015u} and with DeeplabV3+~\cite{chen2018encoder}, which is considered, at the time of writing, the state-of-the-art for image segmentation in computer vision.
Notably, DeeplabV3+ is the segmentation architecture that achieved the best performance in related work on oil spill segmentation in Ref.~\cite{Krestenitis2019b}.

To perform the comparison, we used the Keras implementations of U-net and DeeplabV3+ available at two popular public repositories\footnote{U-net: \url{https://github.com/zhixuhao/unet}, DeeplabV3+: \url{https://github.com/bonlime/keras-deeplab-v3-plus}}.
For this experiment, we used the larger OFCN(32) architecture with configuration C1 described in Tab.~\ref{tab:crossval}.
The DeeplabV3+ is configured with the Xception backbone~\cite{chollet2017xception}.
All the settings are the same as in the previous experiment, with the exception that the models are trained for 400 epochs and the batch size is 16.

\bgroup
\def\arraystretch{1} 
\begin{table}[!ht]
\small
\centering
\begin{tabular}{lccccc}
\cmidrule[1.5pt]{1-6}
\textbf{Model} & \textbf{\# Params.} & \textbf{Tr time (hours)} & \textbf{Tr Acc.} & \textbf{Tr Loss} & \textbf{Val F1 ($\mathcal{D}_1$)} \\
\cmidrule[.5pt]{1-6}
U-net       & 7,760,069     & \textbf{10.1} & 0.984 & 0.058 & 0.741 \\
DeepLabV3+  & 41,049,697    & 15.2 & 0.987 & 0.039 & 0.765 \\
OFCN        & 7,873,729     & 10.9 & \textbf{0.988} & \textbf{0.038} & \textbf{0.775} \\
\cmidrule[1.5pt]{1-6}
\end{tabular}
\caption{Comparison with baselines. Reported is the number of trainable parameters, training time for 400 epochs, training accuracy, training loss, and F1 score on the validation. Best results are in bold. Models are trained on an Nvidia RTX 2080.}
\label{tab:baselines}
\end{table}
\egroup

Tab.~\ref{tab:baselines} reports the number of trainable parameters in each architecture, the time (in hours) necessary to complete 400 epochs of training, the final training accuracy and training loss, and the best F1 score obtained on the validation.
First, we notice that DeeplabV3+ has much more trainable parameters compared to OFCN and U-net, which makes its training almost $50\%$ slower than for the other two architecture.
On the other hand, the training times of OFCN and U-net are comparable.
DeeplabV3+ outperformed U-net but, despite its larger capacity, did not achieve a better performance than the proposed OFCN architecture.

Importantly, we report that in some runs the U-net did not manage to learn anything: the loss was not decreasing and the predicted output was a mask of all zeros for each image in the training set. 
This indicates a strong sensitivity to initialization and a lack of robustness in the U-net model.
Finally, we also experimented with DeeplabV3+ configured with the Mobilenet~\cite{howard2017mobilenets} backbone but we obtained unsatisfactory performance. 

The training graphs showing the evolution of the loss on the training set and the F1 score on the validation set are in Appendix~\ref{sec:additional_details} and they show that none of the models overfits the training data.
 
\subsubsection{Training on the large dataset} 
First, we trained three OFCN(32) models configured with the best hyperparameters settings (C1, C2, C3 reported in Tab.~\ref{tab:crossval}) for 400 epochs.
Training each model on $\mathcal{D}_2$ takes up to one week on an Nvidia RTX 2080.
The results reported in Tab.~\ref{tab:validation} shows that, even in this case, OFCN configured with C1 obtains the best performance: highest accuracy and lowest loss on the training set, highest F1 score on the validation set.

\bgroup
\def\arraystretch{1} 
\begin{table}[!ht]
\small
\centering
\begin{tabular}{lccccc}
\cmidrule[1.5pt]{1-6}
\textbf{ID} & \textbf{Epochs} & \textbf{Time (days)} & \textbf{Tr Acc.} & \textbf{Tr loss} & \textbf{Val F1 ($\mathcal{D}_2$)} \\
\cmidrule[.5pt]{1-6}
C1              & 400 & 6.8 & 0.995 & 0.016 & 0.857 \\
C2              & 400 & 6.9 & 0.990 & 0.047 & 0.750 \\
C3              & 400 & 6.2 & 0.993 & 0.018 & 0.802 \\
C1-2ST        & 400 + 400 & 9.4 & 0.996 & 0.014 & 0.861 \\
\cmidrule[.5pt]{1-6}
C1-2ST-Long & 500 + 3,000 & 54.3 & 0.997 & 0.009 & \textbf{0.892} \\ 
\cmidrule[1.5pt]{1-6}
\end{tabular}
\caption{Validation performance, obtained on $\mathcal{D}_2$ using the three best configurations C1, C2, and C3, the two-step training (2ST) strategy with configuration C1, and a long training of 3000 epochs (Long). We also report the training time, the accuracy, and loss achieved on the training set.}
\label{tab:validation}
\end{table}
\egroup

We also applied the 2-stage training strategy, discussed in Sect.~\ref{sec:training}, using the configuration C1 (C1-2ST in Tab.~\ref{tab:validation}).
The procedure takes approximately $50\%$ extra time since in the first stage the images are down-sampled with a factor of 2 but it yields some performance improvement.

\begin{figure}[!ht]
    \centering
    \subfigure[Training history of C1]{
        \includegraphics[width=0.31\textwidth]{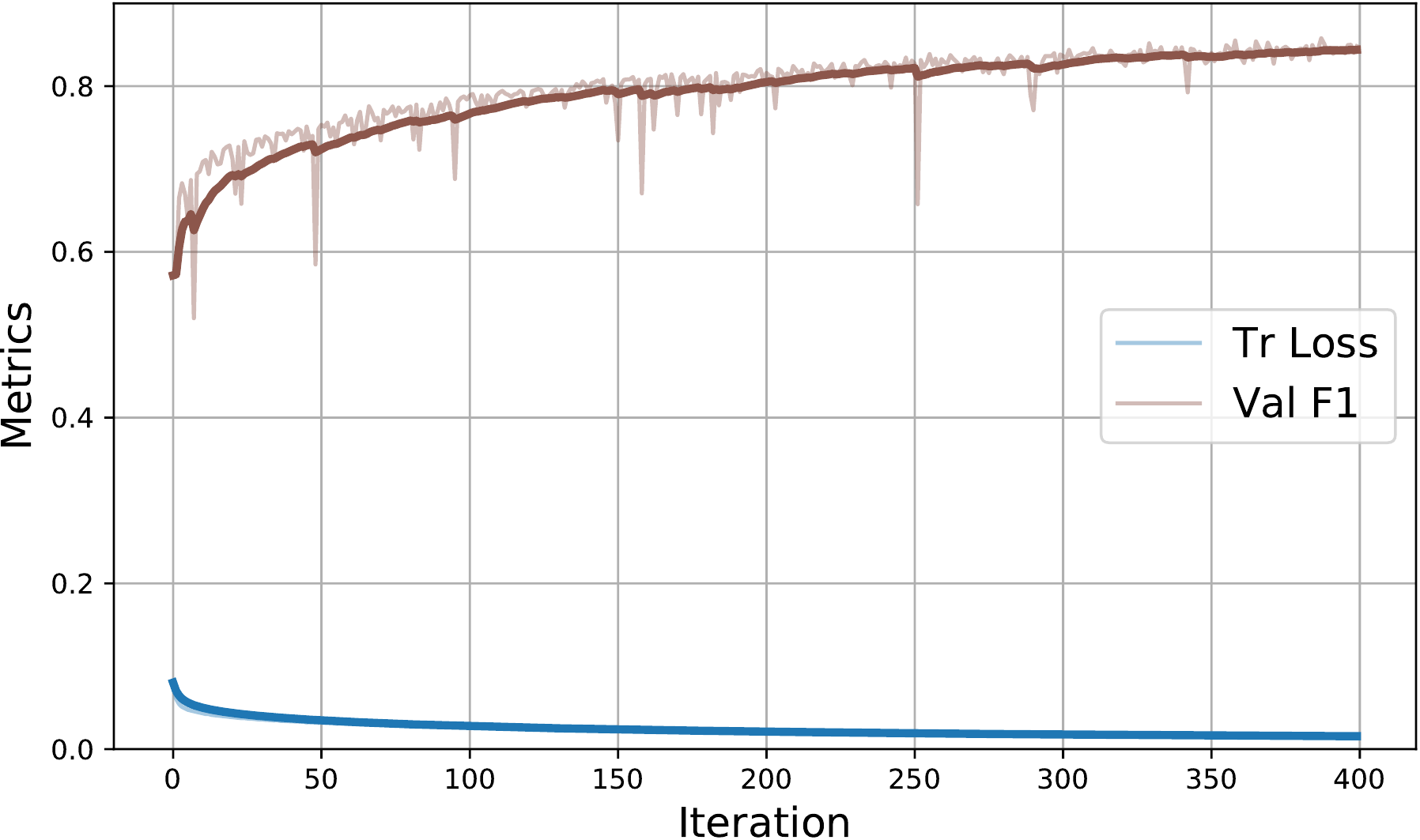}
    }\hspace{-0.5em}%
    ~
    \subfigure[Training history of C2]{
        \includegraphics[width=0.31\textwidth]{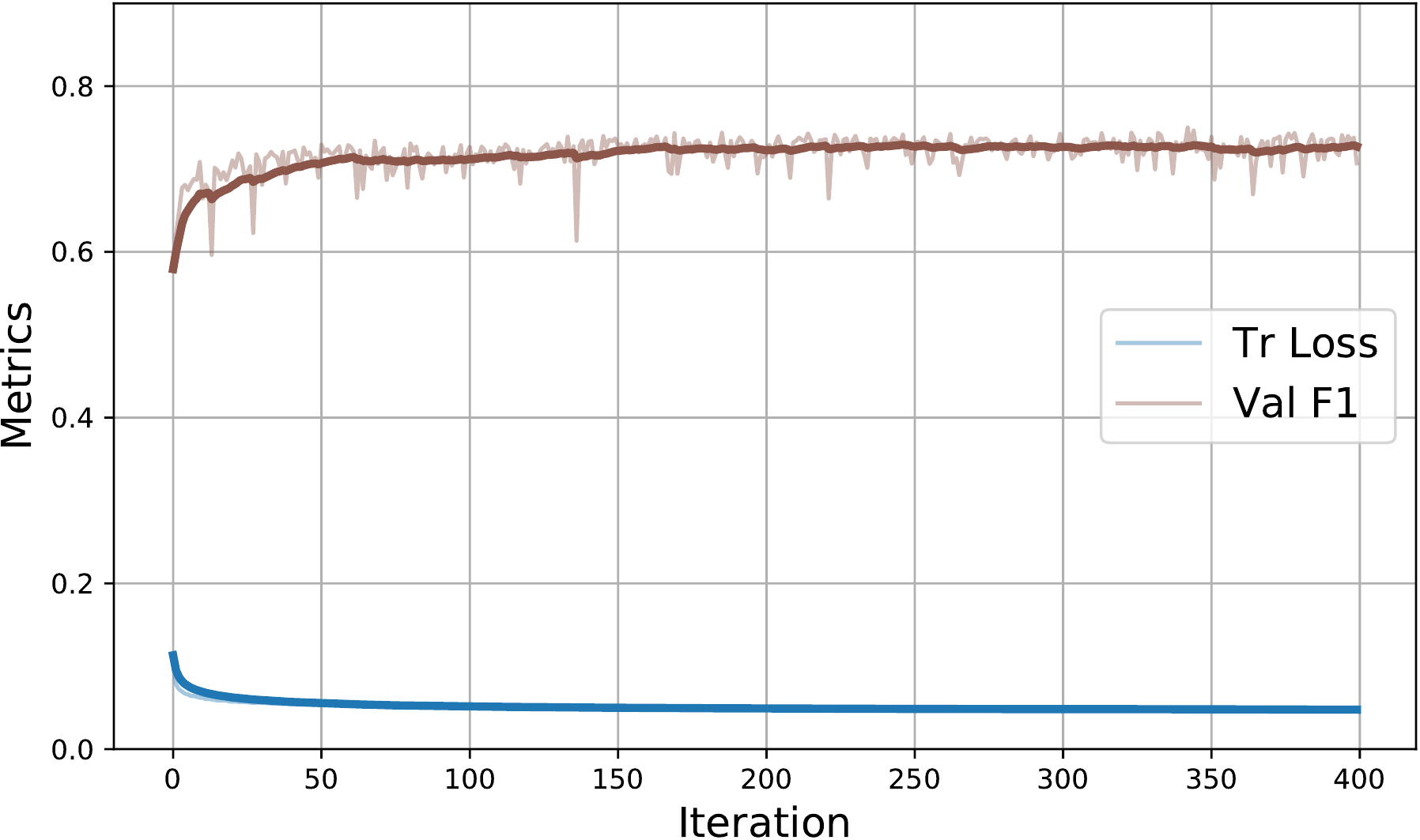}
    }\hspace{-0.5em}%
    ~
    \subfigure[Training history of C3]{
        \includegraphics[width=0.31\textwidth]{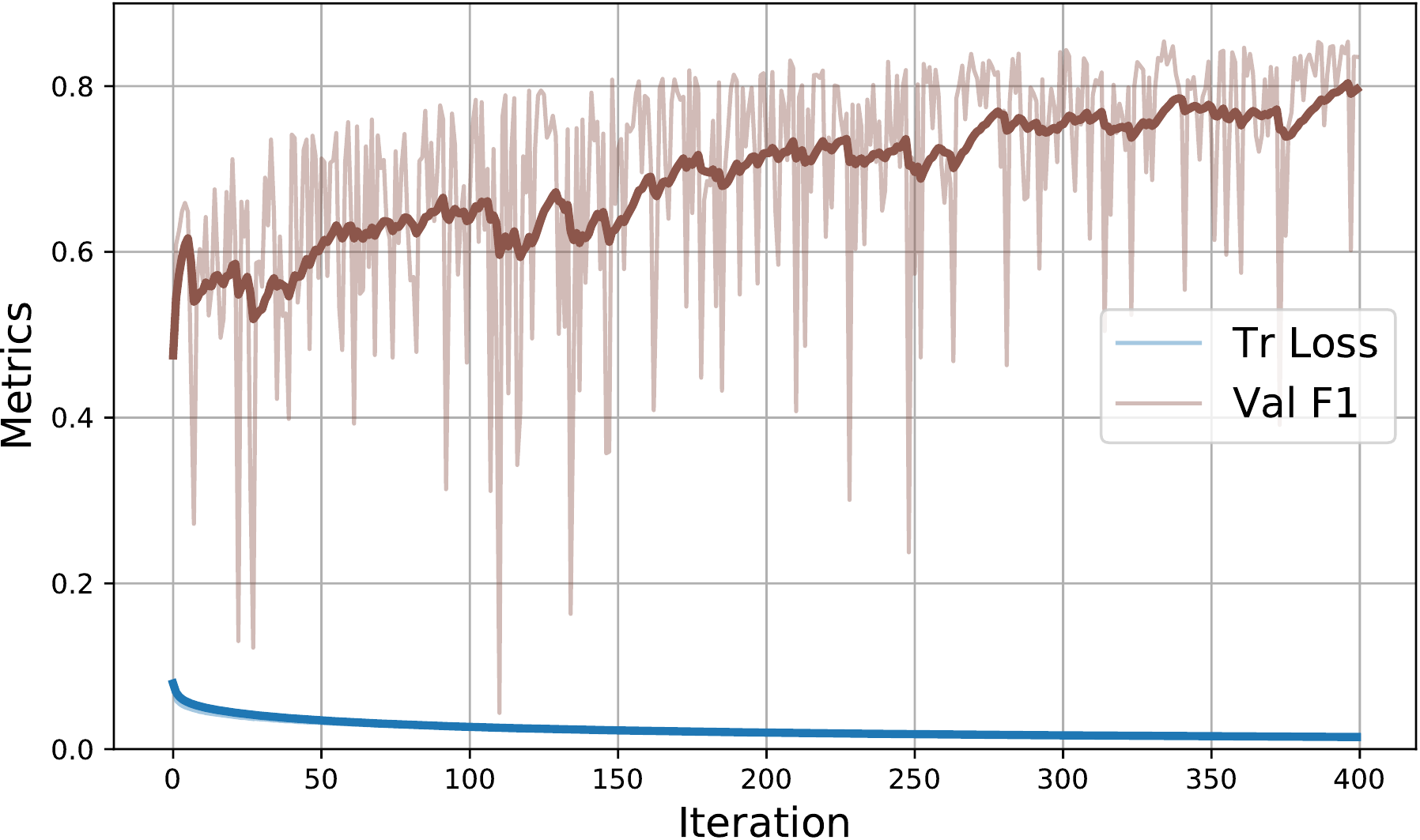}
    }
    \caption{Evolution of training loss and F1 score on validation across the 400 training epochs on dataset $\mathcal{D}_2$. 
    Bold lines indicate a running average with window of size 30.}
    \label{fig:config_loss_plot}
\end{figure}

Fig~\ref{fig:config_loss_plot} depicts the evolution of the training loss and validation F1 score during training for the 3 configurations C1, C2, and C3.
Note that we do not plot the training accuracy, which is always above 98\% from the first epochs, and we did not compute the F1 score on the training set.
The latter requires to compute predictions of the whole training set at each epoch and, given the size of the dataset $\mathcal{D}_2$, it would significantly prolong the training time, which is already in the order of days.

From the plots in Fig~\ref{fig:config_loss_plot}, we notice that the training procedure is much more stable when the OFCN is equipped with the SE module and achieves a higher F1 score when using BN.
Most importantly, none of the models is overfitting on the training set and the F1 score is still improving after 400 epochs.
This suggests that that the training has not converged yet and better performance can be achieved by training the OFCN model for more epochs.

\begin{figure}[!h]
    \centering
    \includegraphics[width=0.8\textwidth]{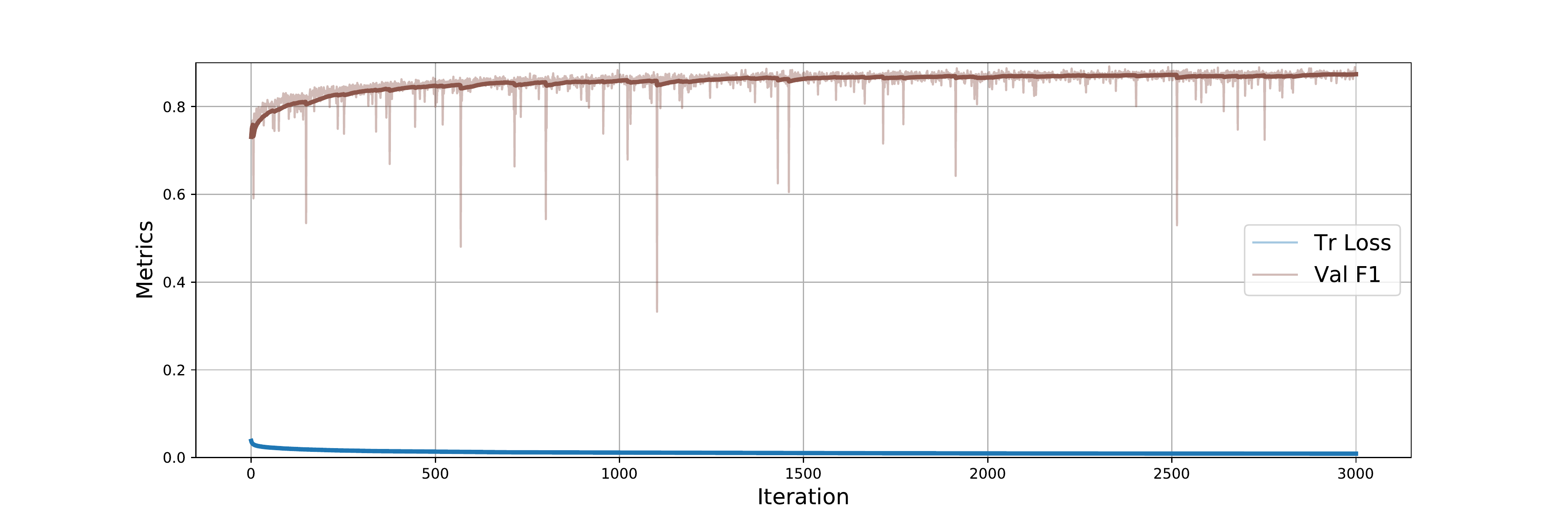}
    \caption{Training history of the model configured with C1 with two-stage training (C1-2ST-Long) on dataset $\mathcal{D}_2$. The plot depicts the evolution of the training loss and F1 score on the validation set over the 3000 epochs in the second stage.
    Bold lines indicate a running average with window of size 30.}
    \label{fig:loss_plot}
\end{figure}

In our last experiment, we trained OFCN(32) configured with C1 for 500 epochs in phase one (down-scaled images) and then 3,000 epochs on full resolution images.
Training this model took almost two months and we obtained a significant improvement, reaching an exceptional 0.892 F1 score on the validation set.
Fig.~\ref{fig:loss_plot} reports the training statistics for the second training phase.
After 3,000 epochs the F1 score has finally stabilized but we also notice that the network is still not overfitting the training set.

\begin{figure}[!h]
    \centering
    \subfigure{
        \includegraphics[width=0.48\columnwidth]{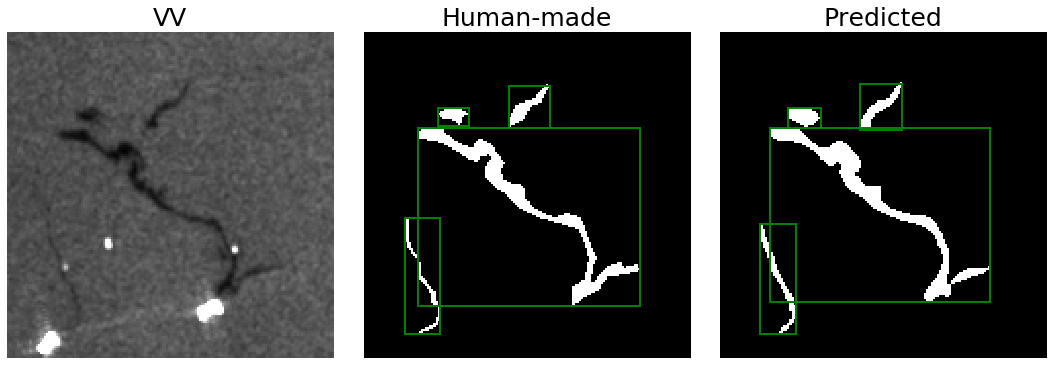}
    }\hspace{-.4cm}%
    ~
    \subfigure{
        \includegraphics[width=0.48\columnwidth]{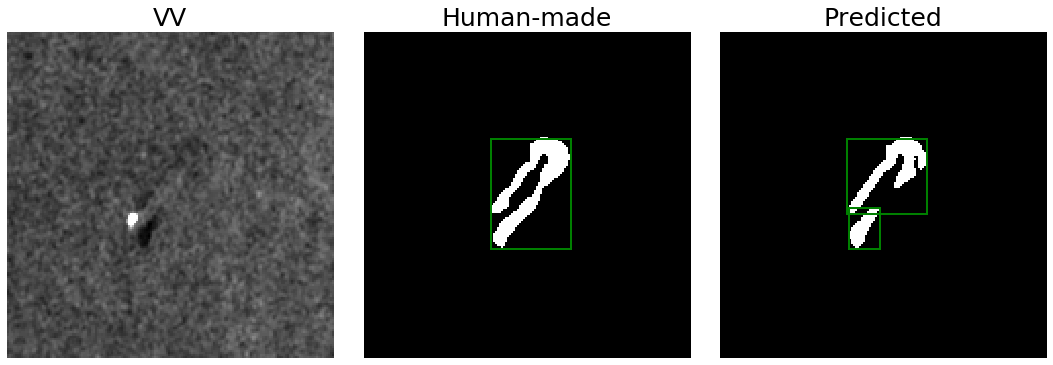}
    }
    
    \subfigure{
        \includegraphics[width=0.48\columnwidth]{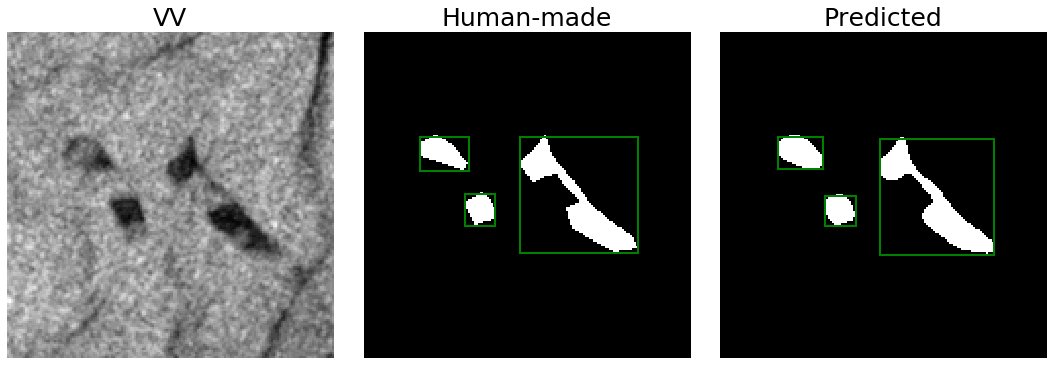}
    }\hspace{-.4cm}%
    ~
    \subfigure{
        \includegraphics[width=0.48\columnwidth]{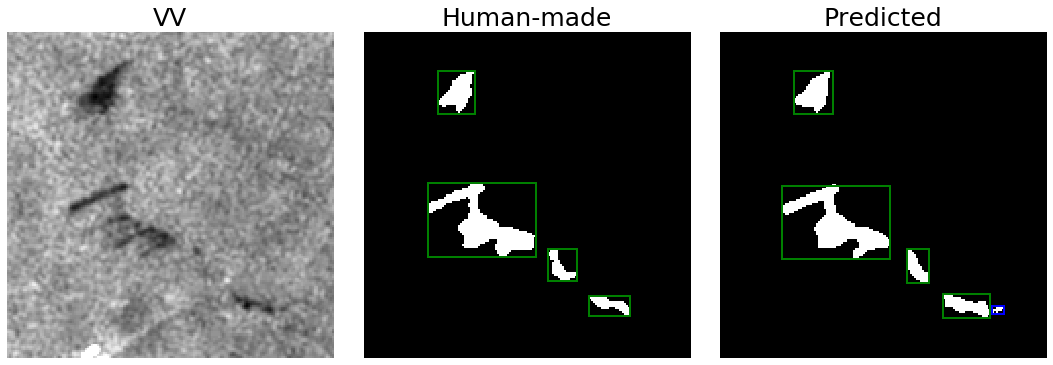}
    }
    
    \subfigure{
        \includegraphics[width=0.48\columnwidth]{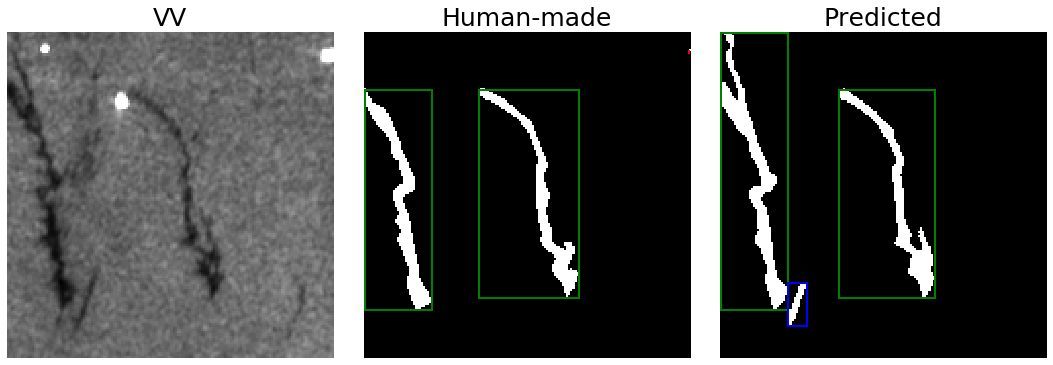}
    }\hspace{-0.4cm}%
    ~
    \subfigure{
        \includegraphics[width=0.48\columnwidth]{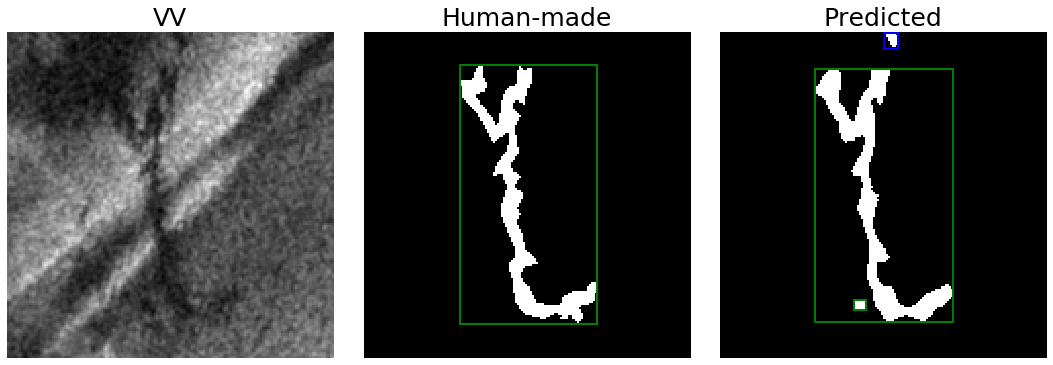}
    }
    
    \subfigure{
        \includegraphics[width=0.48\columnwidth]{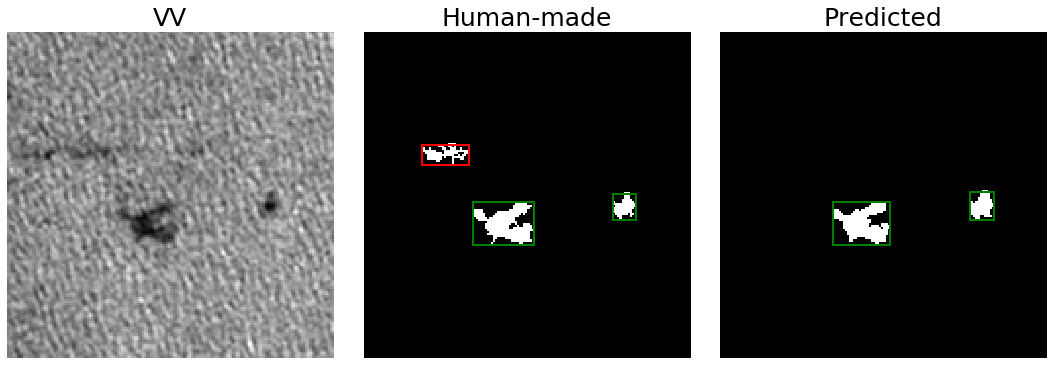}
    }\hspace{-0.4cm}%
    ~
    \subfigure{
        \includegraphics[width=0.48\columnwidth]{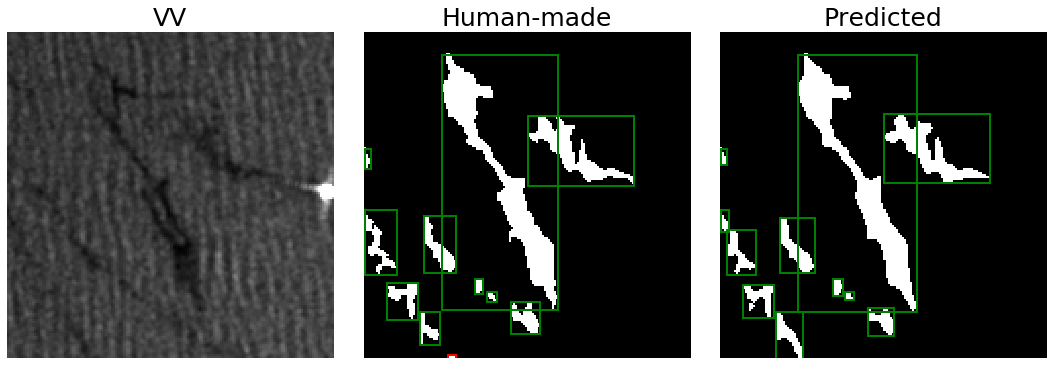}
    }
    \caption{Examples of segmentation masks predicted by the OFCN on the validation set of $\mathcal{D}_2$. From left to right: the VV input channel, the mask made by the human operator, the OFCN output thresholded at 0.5 (values $\leq 0.5 \rightarrow 0$, values $\geq 0.5 \rightarrow 1$).
    Green bounding boxes are TP (the oil spill appears both in the human-made and the predicted mask), blue boxes are FP (the OFCN detects an oil spill that is not present in the human-made mask), and red boxes FN (the oil spill is in the human-made mask but is not detected by the OFCN).}
    \label{fig:D2_pred}
\end{figure}

Fig.~\ref{fig:D2_pred} reports some examples of segmentation masks predicted by the OFCN on the validation set of $\mathcal{D}_2$.
From left to right, we depict the VV input channel, the ground truth mask made by human operator, and the OFCN prediction thresholded at 0.5 (values $\leq 0.5 \rightarrow 0$, values $\geq 0.5 \rightarrow 1$).
To facilitate the interpretation of the results, we generated a bounding box around both the oil spills in the mask generated by the operator and the mask predicted by OFCN.
If two bounding boxes overlap, they are \textit{True Positives} (TP) and are coloured in green.
Bounding boxes appearing only in the predicted mask are \textit{False Positives} (FP) and are depicted in blue.
Finally, \textit{False Negatives} (FN) are the red boxes appearing only in the human-made label and not in the prediction.

FPs are more common as they can arise from small details that might be overlooked by the human operator and often appear on the edge of the oil spill outline.
On the other hand, FNs are very rare meaning that our model misses very few of the human-detected oil spills.
Having a low amount of FNs is particularly important because FP can always be discarded during a post-analysis, whereas a missed detection cannot be recovered. 

From now on, by OFCN we will refer to the OFCN(32) model trained with configuration C1-2ST-Long (see Tab.~\ref{tab:validation}).

\subsubsection{Segmentation performance and incidence angle.}
\begin{figure}[!h]
\centering
    \includegraphics[width=0.6\columnwidth]{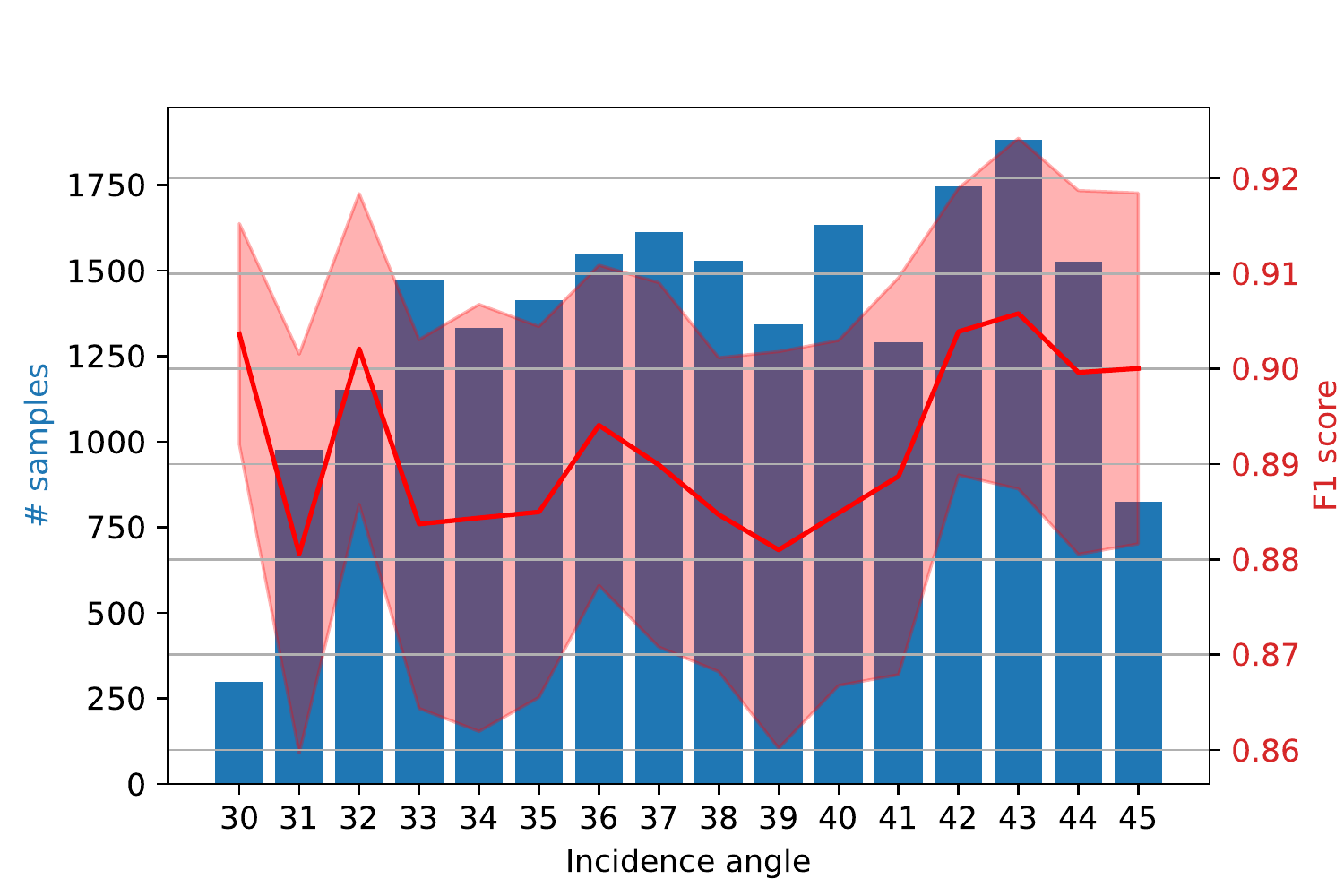}
    \caption{Variation of the F1 score on the validation set according to the incidence angle of the satellite.}
    \label{fig:f1_vs_lia}
\end{figure}

Here, we investigate how much the incidence angle of the satellite affects oil spills detection.
All the patches in our dataset are characterized by an incidence angle between 30 and 45 degrees, which is the range where the scanSAR mode of Sentinel-1 is operating~\cite{sentinel1}.
Our initial hypothesis was that oil spills are detected more easily at medium inclinations since high and low incidence angles yield low oil-sea contrast \cite{ALPERS2017133,Gade1998}.
However, the results disproved our hypothesis. 
In Fig.~\ref{fig:f1_vs_lia}, the red line shows the mean detection F1 score obtained for oil spills at a given incidence angle; the red area shows the standard deviation; the blue bars indicate the number of oil spills for each incidence angle.
To assess if there are statistical differences between the F1 scores obtained at different incidence angles, we perform the Kruskal-Wallis H-test, which is a non-parametric version of ANOVA that makes only a few assumptions about the characteristics of the population from which the data originate~\cite{landau2004handbook}.
With the Kruskal-Wallis, we test the null hypothesis that the population median of all of the groups are equal and we obtain H-statistic = 54.14 and $p$-value $< 10^{-4}$. 
Since the $p$-value is much lower than 0.05, the null hypothesis cannot be rejected, meaning that there is not a statistically significant difference between the F1 scores obtained at different incidence angles.

However, we recall that the incidence angles available are in the range 30 to 45 degrees, where oil spill detection is usually preferred and Bragg scattering dominates \cite{holt2004a,ALPERS2017133}. 
Outside this range, specular reflection from both oil slicks and clean sea occurs resulting in lower oil-sea contrast~\cite{ALPERS2017133,Minchew2012b}. 
It is also interesting to notice fewer oil samples at the near and far incidence angles (30 and 45 degrees), where oil slick detection is challenging.

\subsubsection{Visualization of the learned filters}

\begin{figure}[!ht]
    \centering
    \includegraphics[keepaspectratio,width=.65\textwidth]{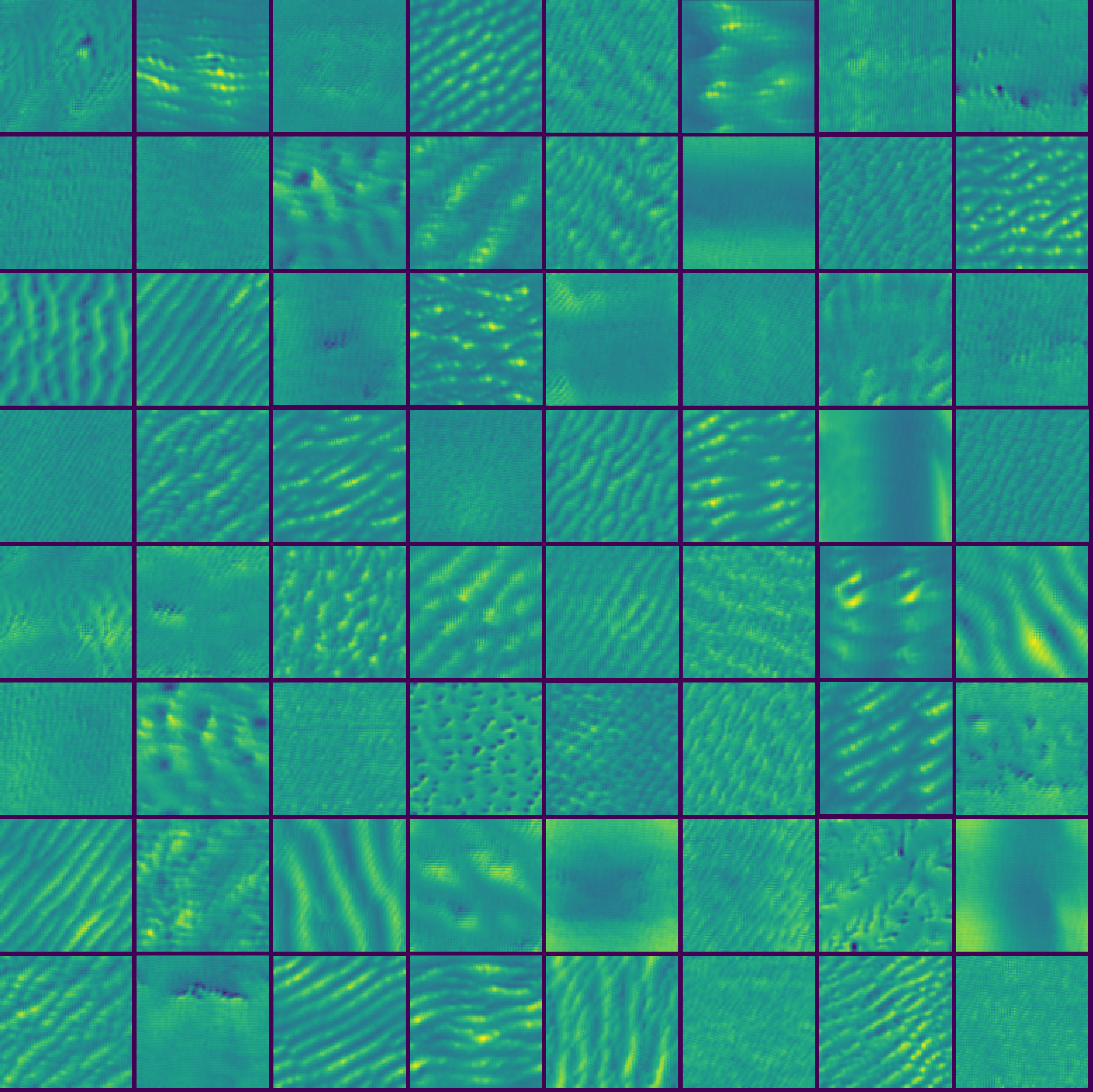}    
    \caption{Filters visualization. We synthetically generated the input images that maximally activate 64 of the 512 filters in the first convolutional layer of the ``Enc Block 512'' at the bottom of the OFCN.}
    \label{fig:filters}
\end{figure}

To provide an interpretation of what the OFCN is learning, we synthetically generated images with patterns that maximally activate the convolutional filters in OFCN.
To generate the images, we rely on the Activation Maximization technique~\cite{nguyen2016synthesizing}: we start from an input image containing random noise and we gradually update it by ascending the gradient $\frac{\delta Activation Maximization Loss}{\delta input}$.

Fig.~\ref{fig:filters} depicts the patterns that maximally activate the first 64 of the 512 filters in the first convolutional layer of the ``Enc Block 512'' (the one at the bottom of OFCN in Fig.~\ref{fig:unet}).
Interestingly, it is possible to notice several wave patterns, meaning that the OFCN is looking for waves when trying to detect the oil spills. 
This is reasonable since ocean surface waves are present in moderate wind conditions, which are the most favourable for detecting oil spills \cite{alpers1989,Singh1986}. 
In fact, oil spills cannot be detected if the sea surface is too flat due to low wind or is too irregular due to high wind. For low winds, the backscatter response is similar between the calm ocean surface and oil slicks~\cite{Girard2005}. Whereas for strong wind, the oil can break and/or sink due to upper surface layer turbulence. 
The wind speed range for optimal oil-sea contrast is suggested to be 2-3m/s to 10-14m/s \cite{alpers1989,Singh1986}.

\subsubsection{Results on the test set $\mathcal{D}_\text{test}$}

\bgroup
\def\arraystretch{1} 
\setlength\tabcolsep{1em} 
\begin{table}[!ht]
\small
\centering
\begin{tabular}{lccccc}
\cmidrule[1.5pt]{1-6}
\textbf{ID} & \textbf{F1} & \textbf{IoU} & \textbf{TP} & \textbf{FP} & \textbf{FN} \\
\cmidrule[.5pt]{1-6}
T1 & 0.73 & 0.81 & 2  & 7  & 0 \\
T2 & 0.44 & 0.36 & 11 & 59 & 0 \\
T3 & 0.83 & 0.52 & 31 & 14 & 5 \\
\cmidrule[1.5pt]{1-6}
\end{tabular}
\caption{Performance obtained on the 3 SAR products in $\mathcal{D}_\text{test}$ used as test set.}
\label{tab:test}
\end{table}
\egroup

Tab.~\ref{tab:test} reports the performance obtained on the three test products of $\mathcal{D}_\text{test}$, whose details are in Tab.~\ref{tab:test_data}.
Compared to the validation set, the proportion of oil pixels is much lower in the three SAR products. 
Also, given the large size of the images that extend up to 860km, there is a higher chance of detecting FPs.
For these reasons, the F1 scores Tab.~\ref{tab:test} are lower compared to the score obtained on the validation set (0.892).
We note, however, that a large portion of FPs is due to small bounding boxes not marked by the human operators.

\begin{figure}[!h]
    \centering
    \subfigure[T1 hits]{
        \includegraphics[height=4cm, width=5.5cm]{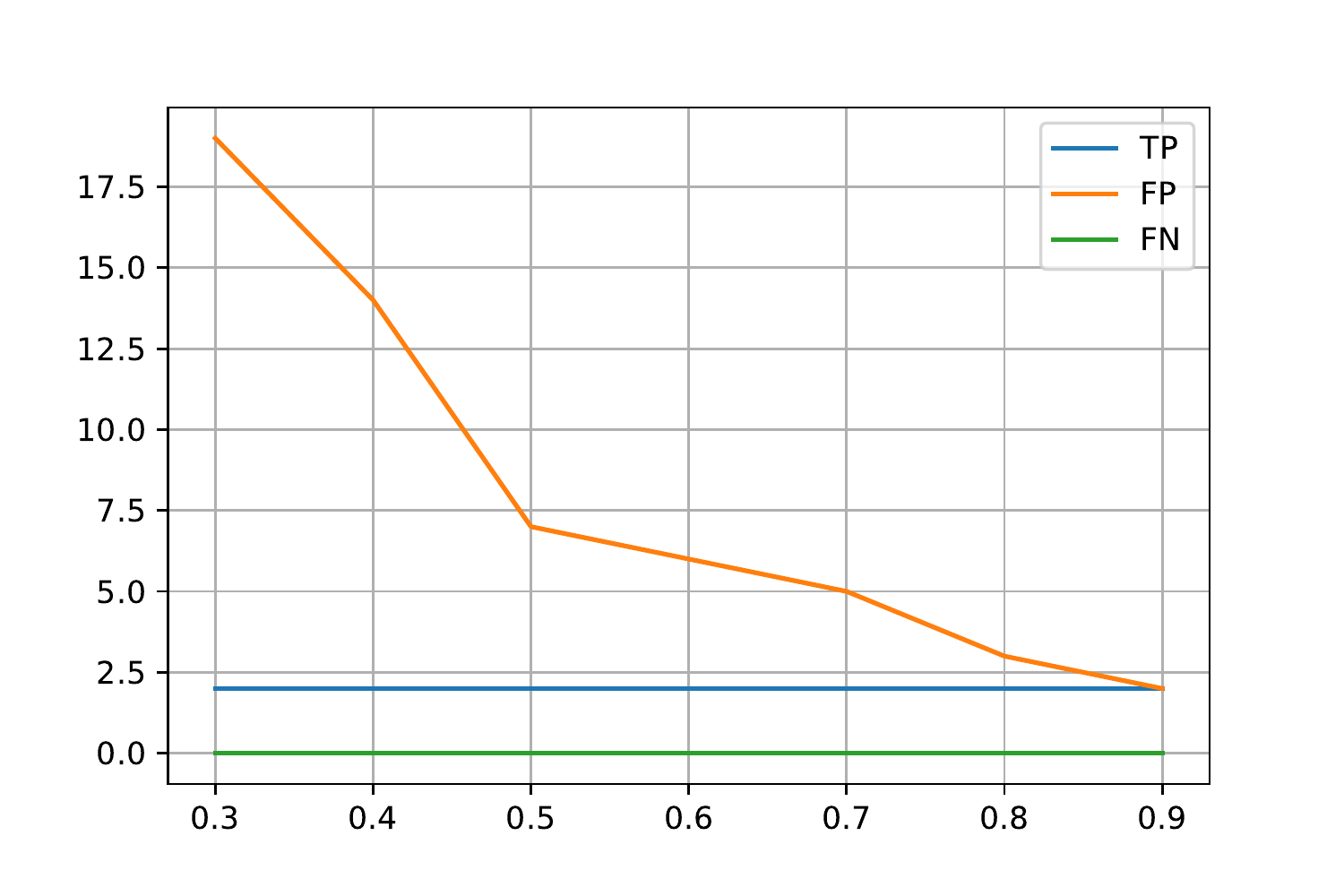}
    }\hspace{-1cm}%
    ~
    \subfigure[T2 hits]{
        \includegraphics[height=4cm, width=5.5cm]{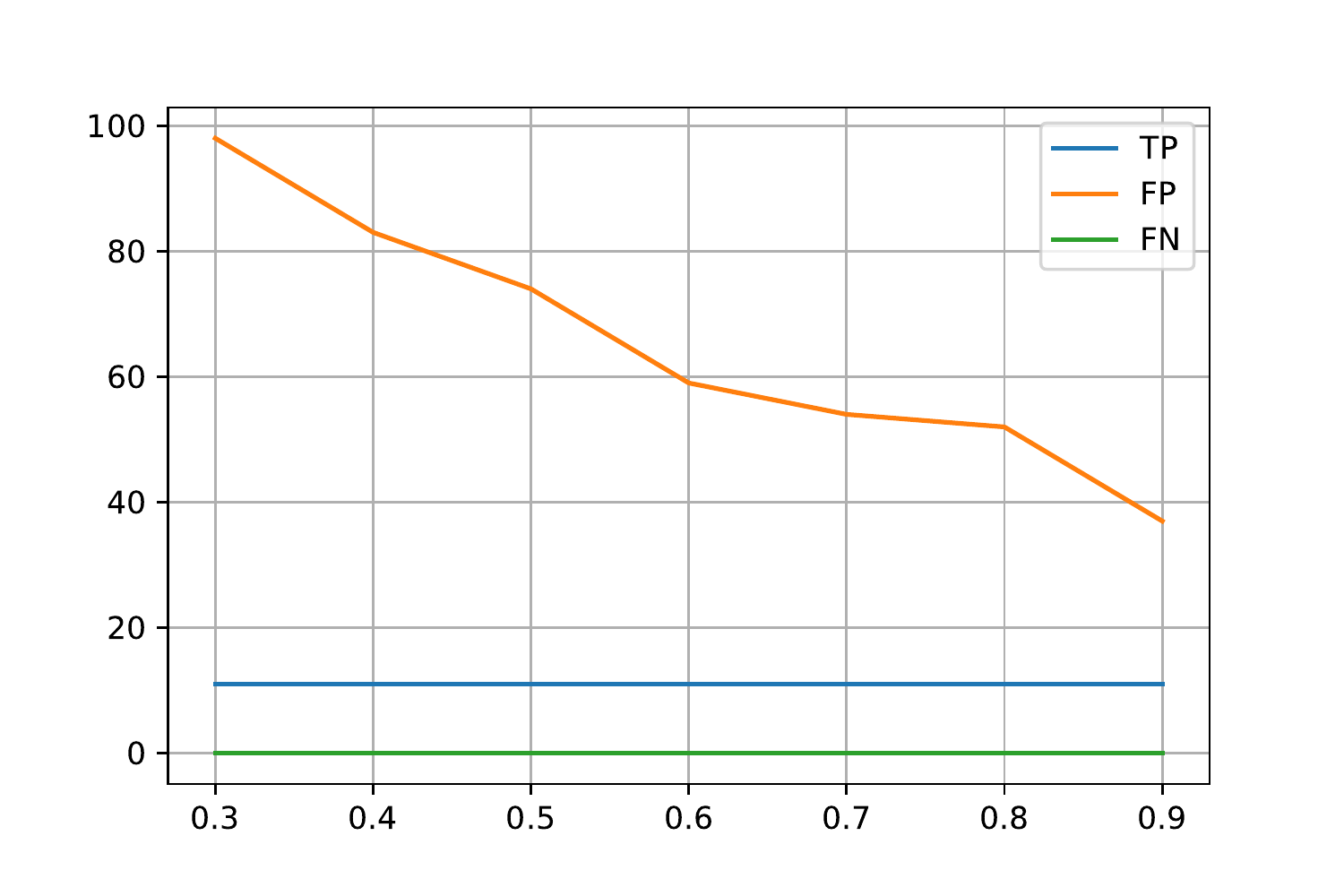}
    }\hspace{-1cm}%
    ~
    \subfigure[T3 hits]{
        \includegraphics[height=4cm, width=5.5cm]{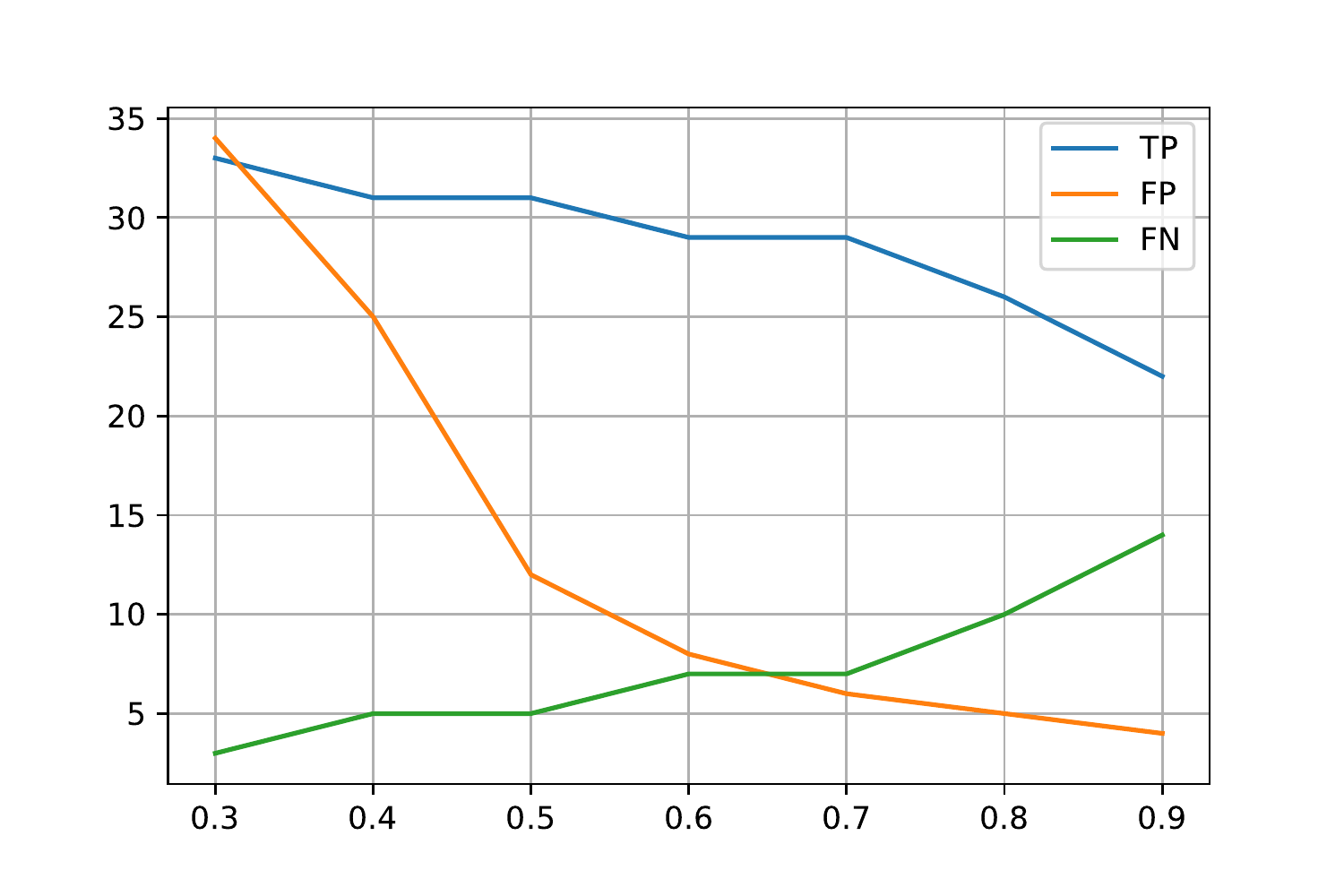}
    }
    
    \subfigure[T1 metrics]{
        \includegraphics[height=4cm, width=5.5cm]{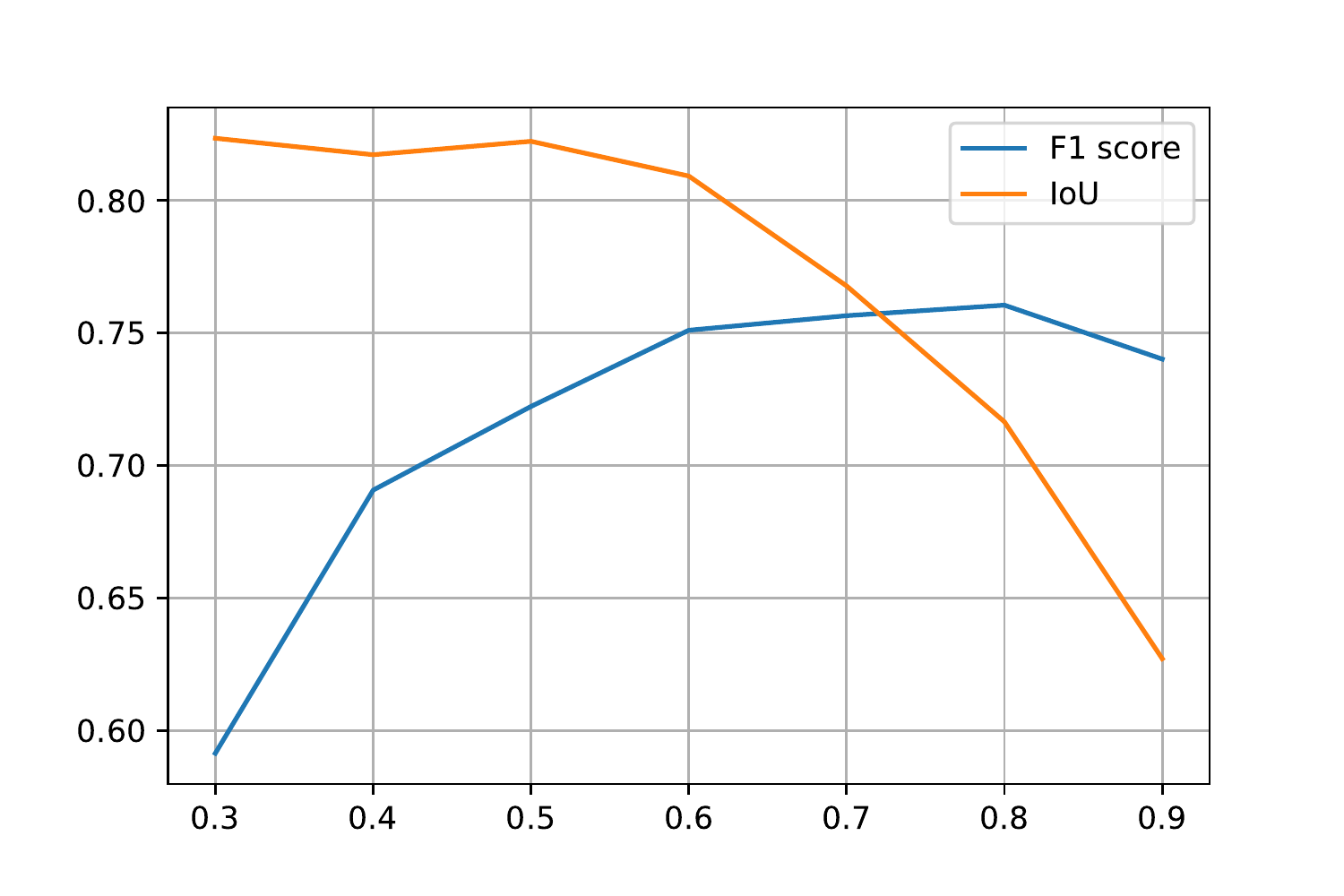}
    }\hspace{-1cm}%
    ~
    \subfigure[T2 metrics]{
        \includegraphics[height=4cm, width=5.5cm]{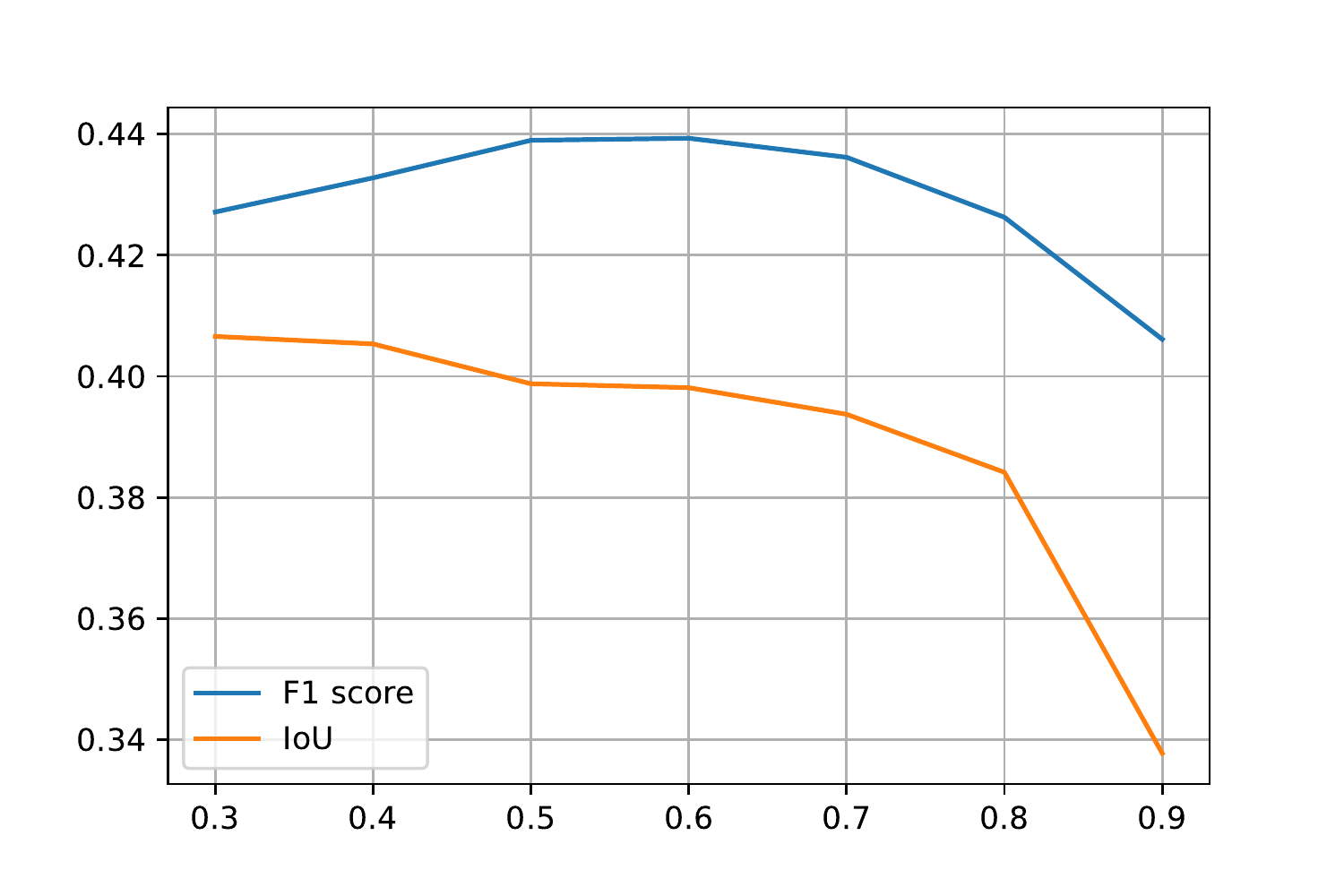}
    }\hspace{-1cm}%
    ~
    \subfigure[T3 metrics]{
        \includegraphics[height=4cm, width=5.5cm]{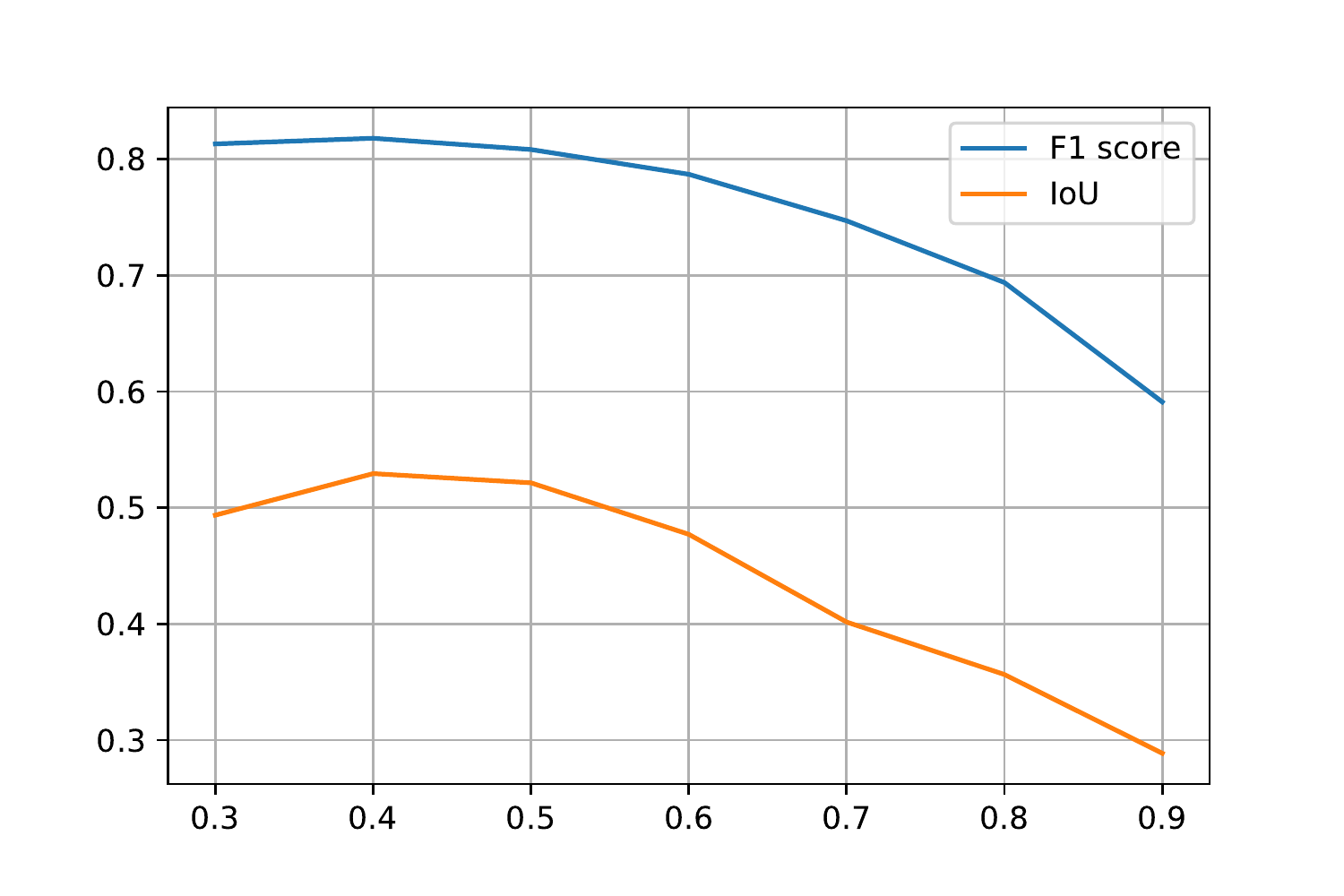}
    }
    \caption{The x-axis always denotes the value of the rounding threshold $\tau$. 
    (a-c) number of True Positive (TP), False Positive (FP), and False Negative (FN) detection obtained by using a different threshold $\tau$ on the soft output of the OFCN. 
    (d-e) values of F1 score and IoU for different $\tau$.}
    \label{fig:test_thresh}
\end{figure}

\begin{figure}[!ht]
    \centering
    \subfigure[SAR input]{
        \includegraphics[width=0.3\textwidth]{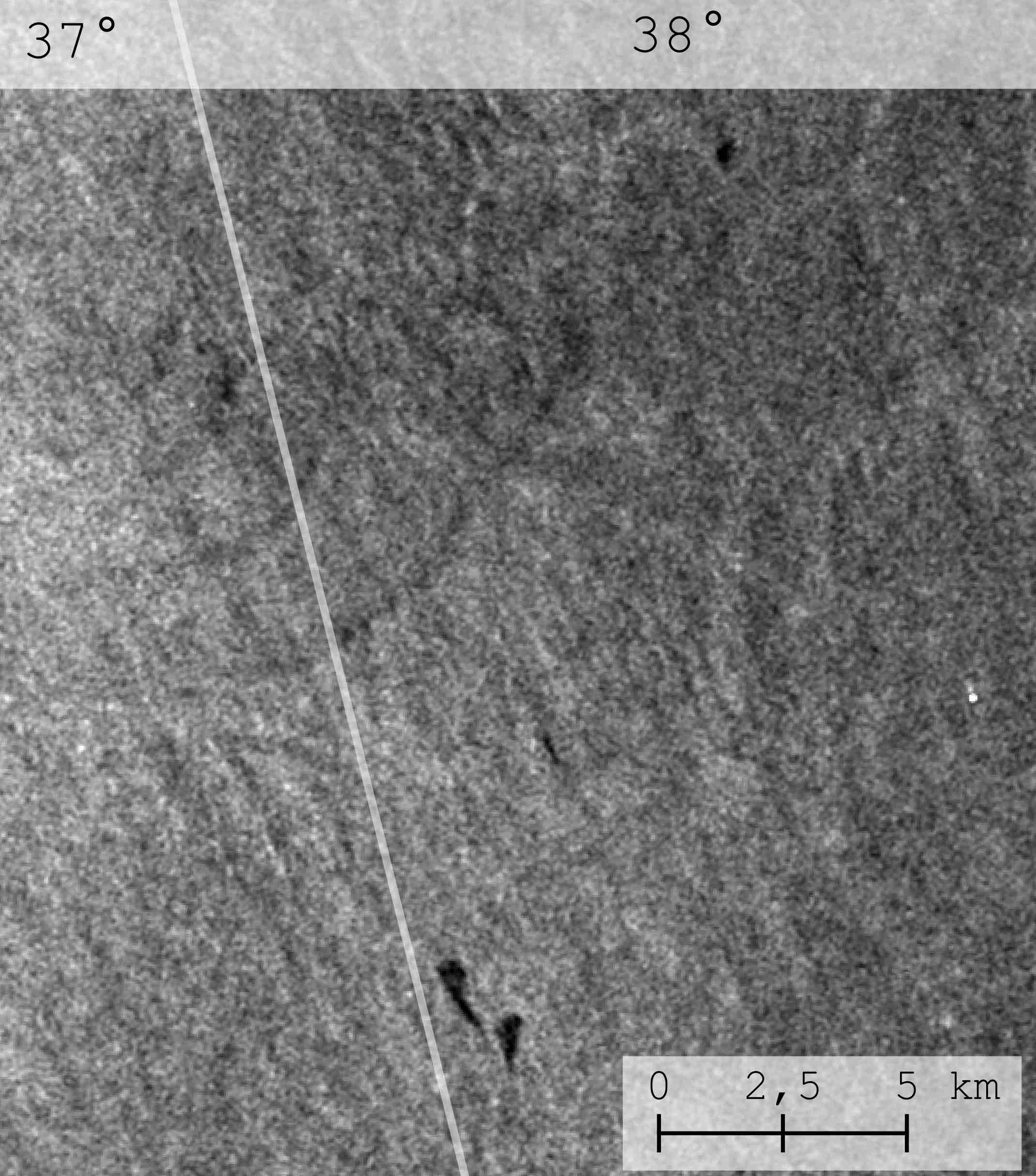}
    }\hspace{-0.7em}%
    ~
    \subfigure[Human-made mask]{
        \includegraphics[width=0.3\textwidth]{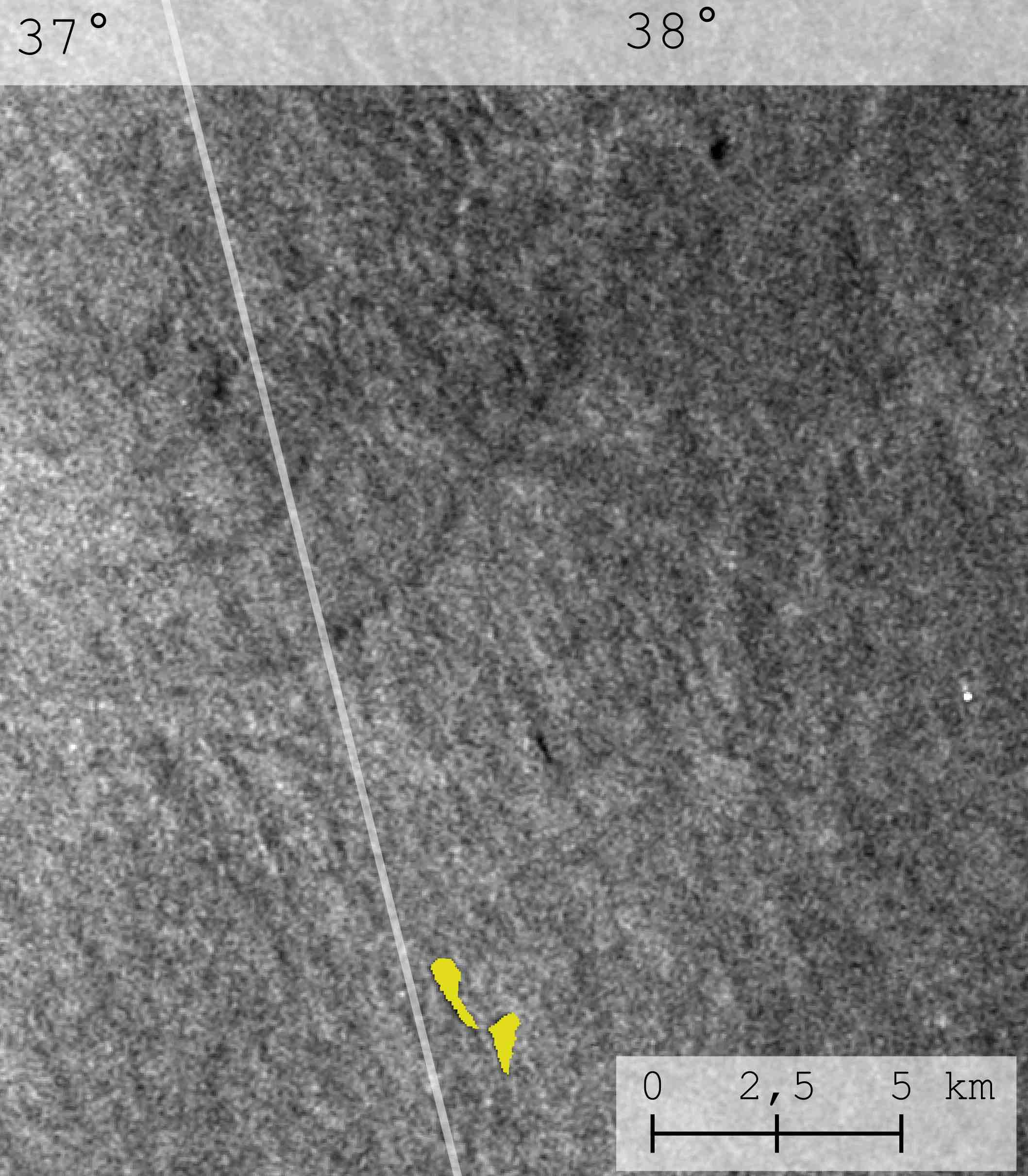}
    }\hspace{-0.7em}%
    ~
    \subfigure[Predicted mask]{
        \includegraphics[width=0.3\textwidth]{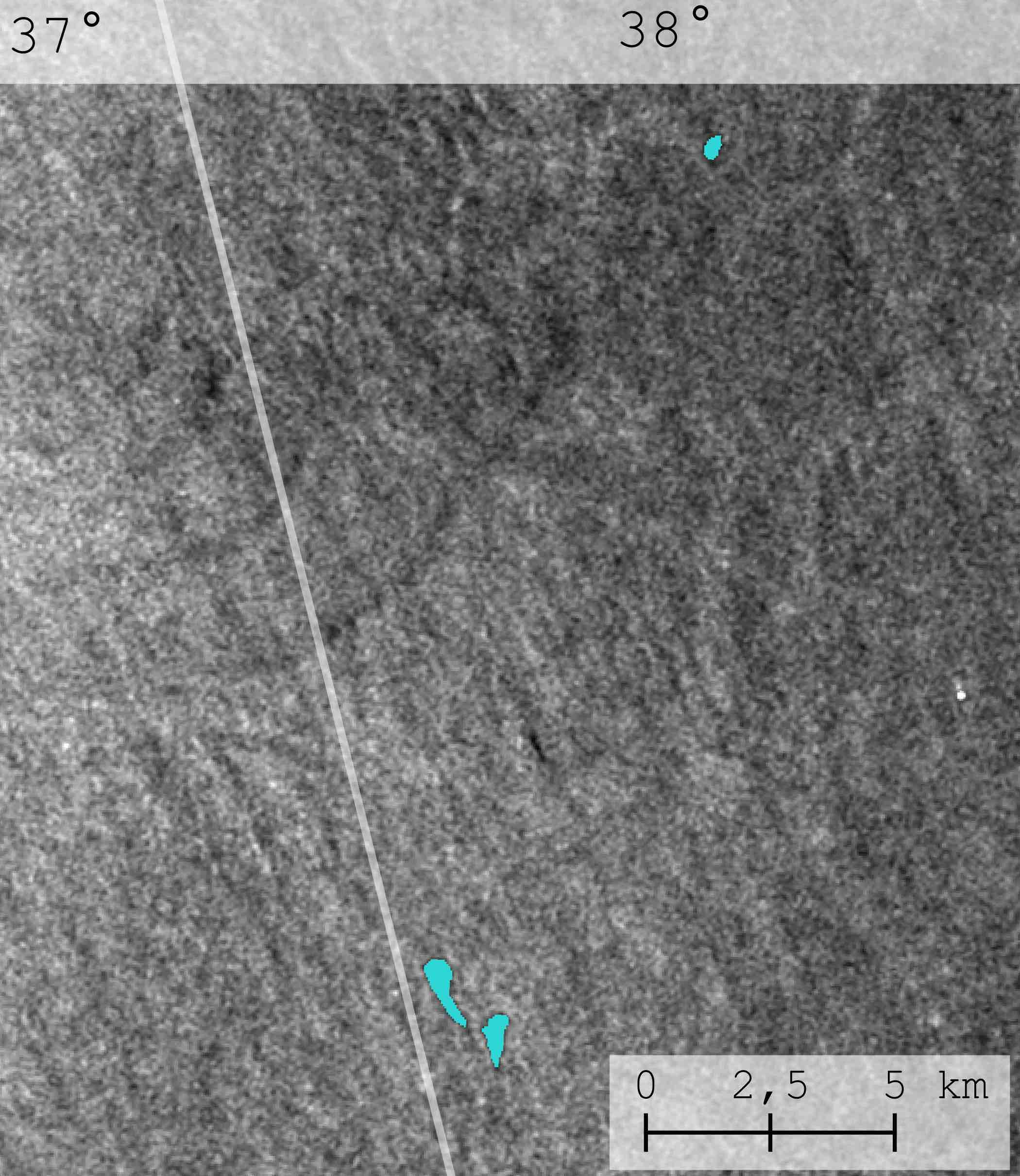}
    }\hspace{-0.7em}%
    
    \subfigure[SAR input]{
        \includegraphics[width=0.3\textwidth]{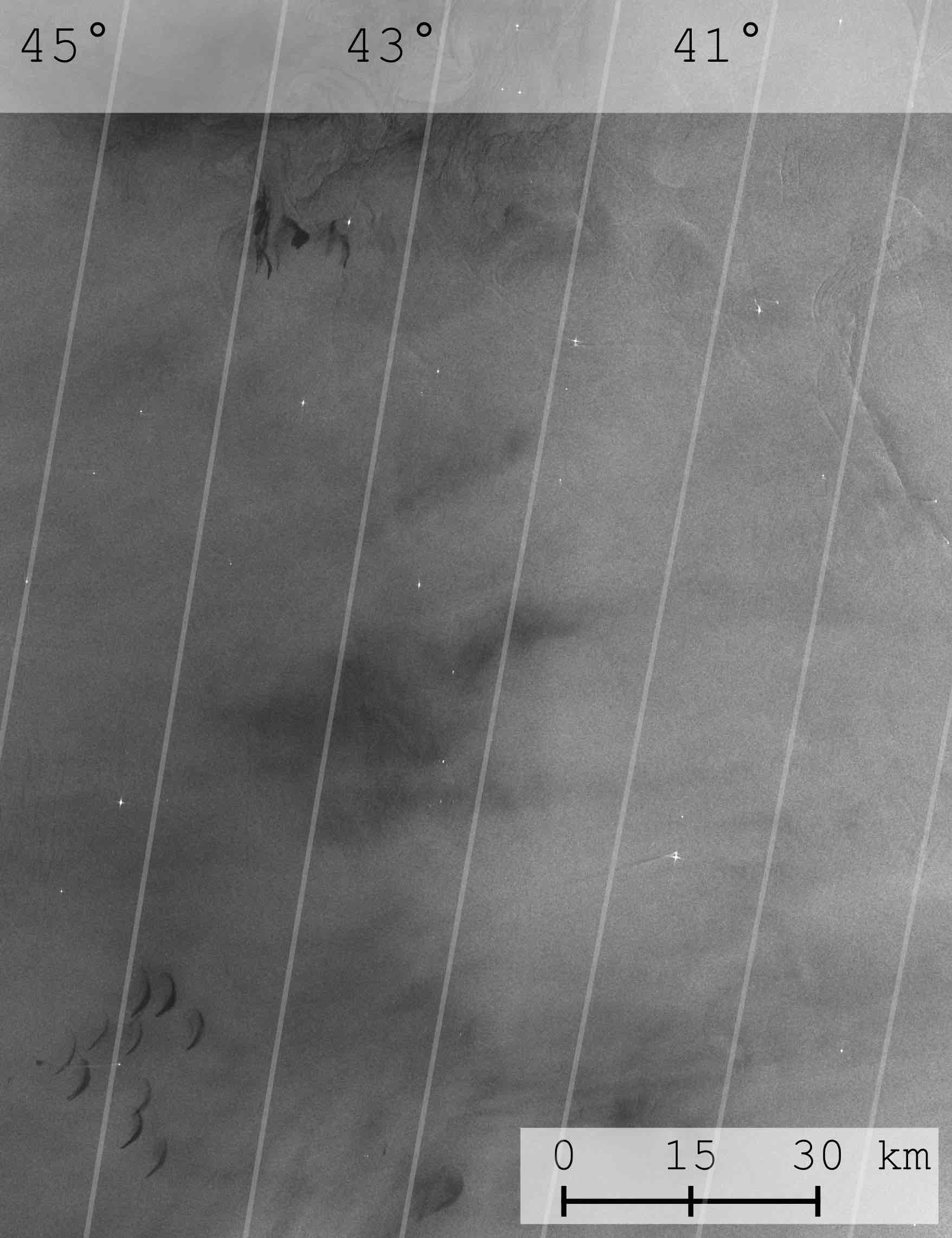}
    }\hspace{-0.7em}%
    ~
    \subfigure[Human-made mask]{
        \includegraphics[width=0.3\textwidth]{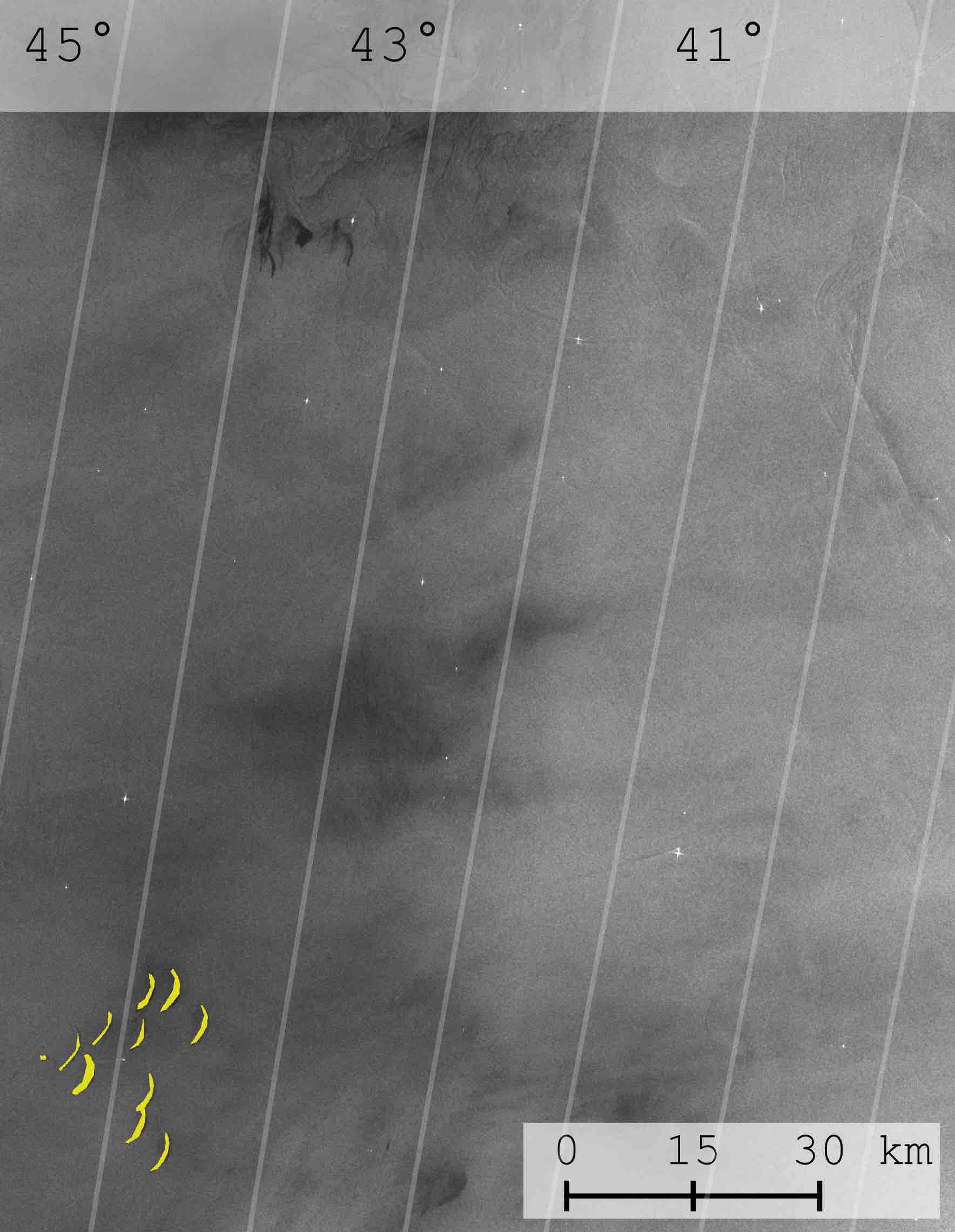}
    }\hspace{-0.7em}%
    ~
    \subfigure[Predicted mask]{
        \includegraphics[width=0.3\textwidth]{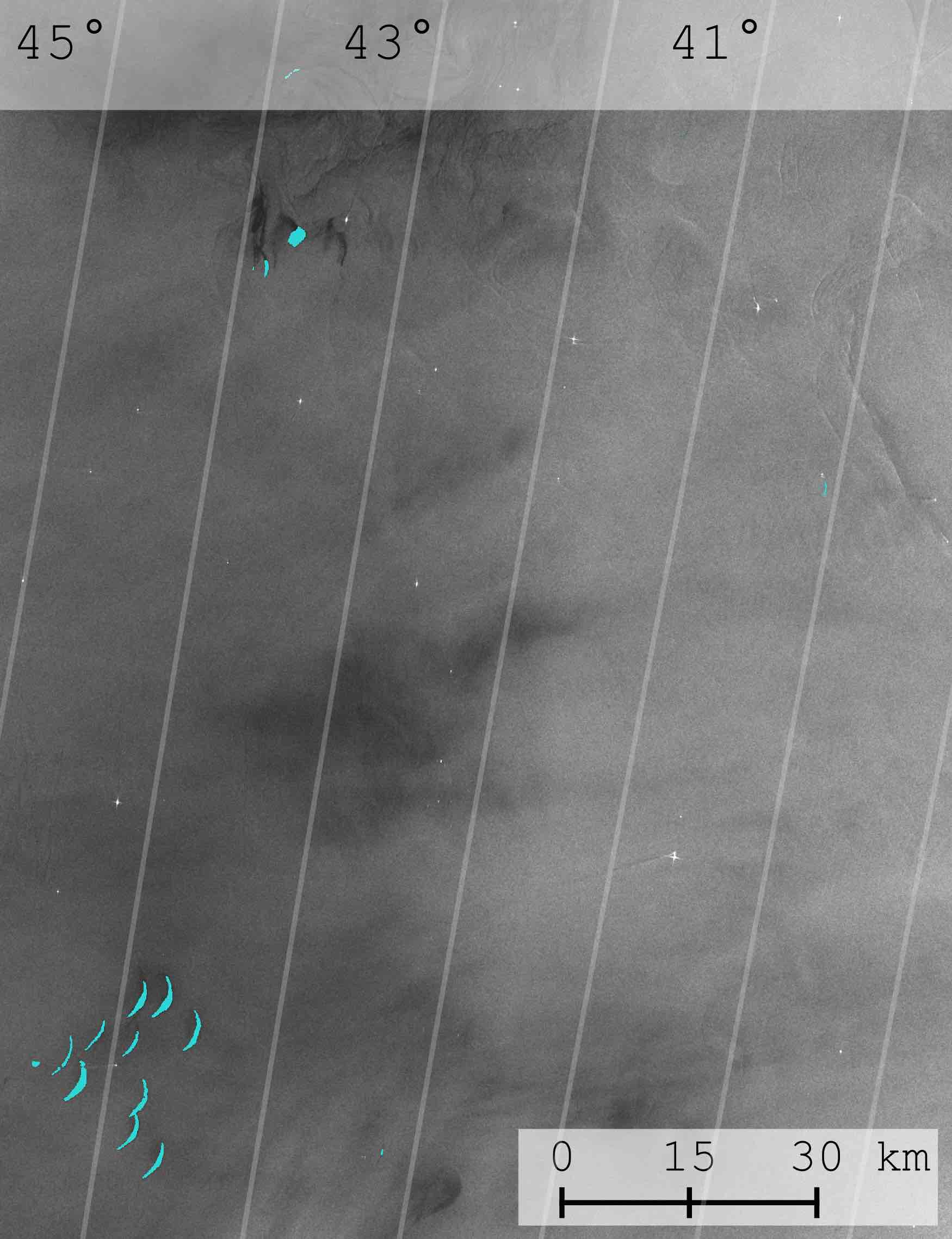}
    }\hspace{-0.7em}%
    
    \subfigure[SAR input]{
        \includegraphics[width=0.3\textwidth]{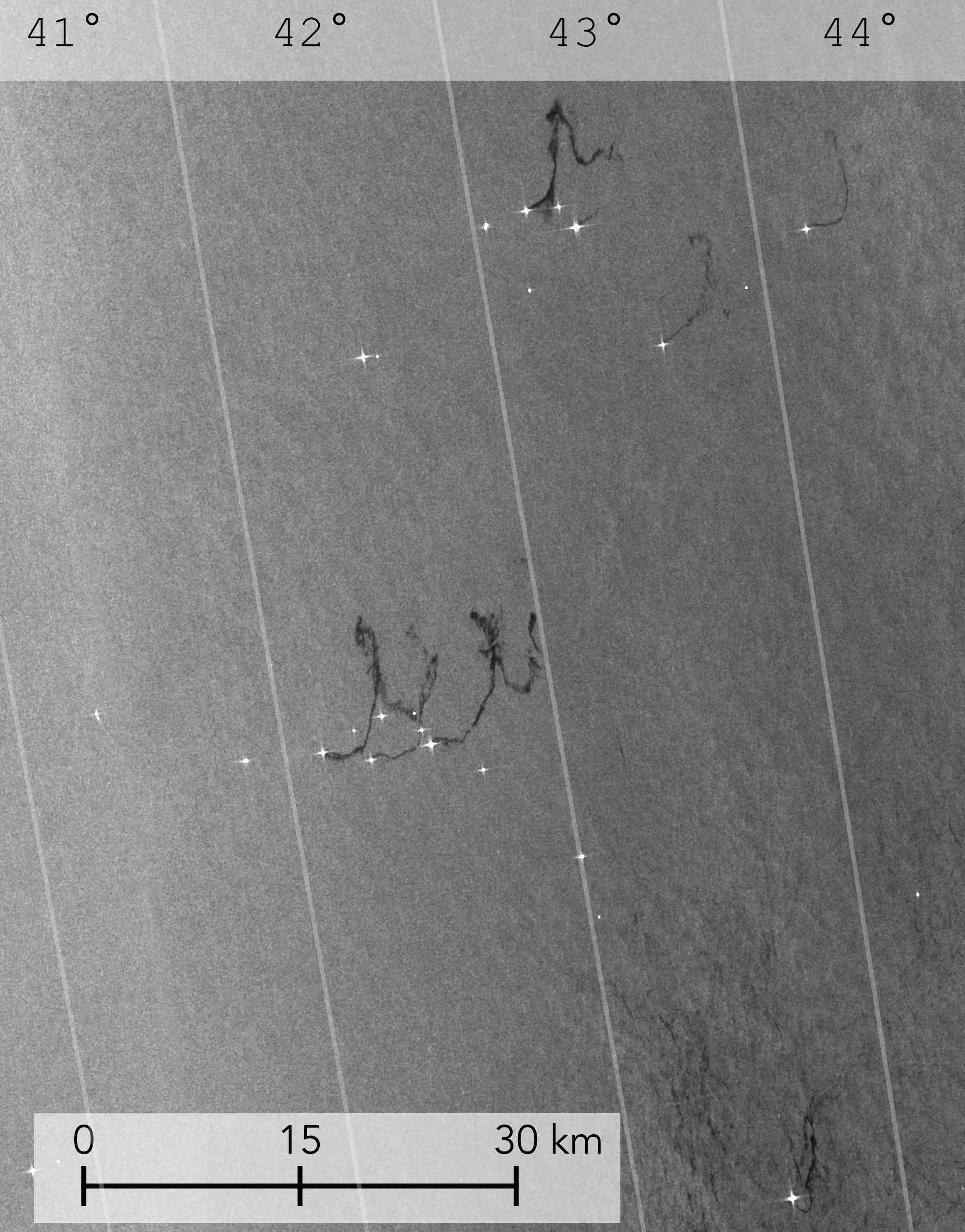}
    }\hspace{-0.7em}%
    ~
    \subfigure[Human-made mask]{
        \includegraphics[width=0.3\textwidth]{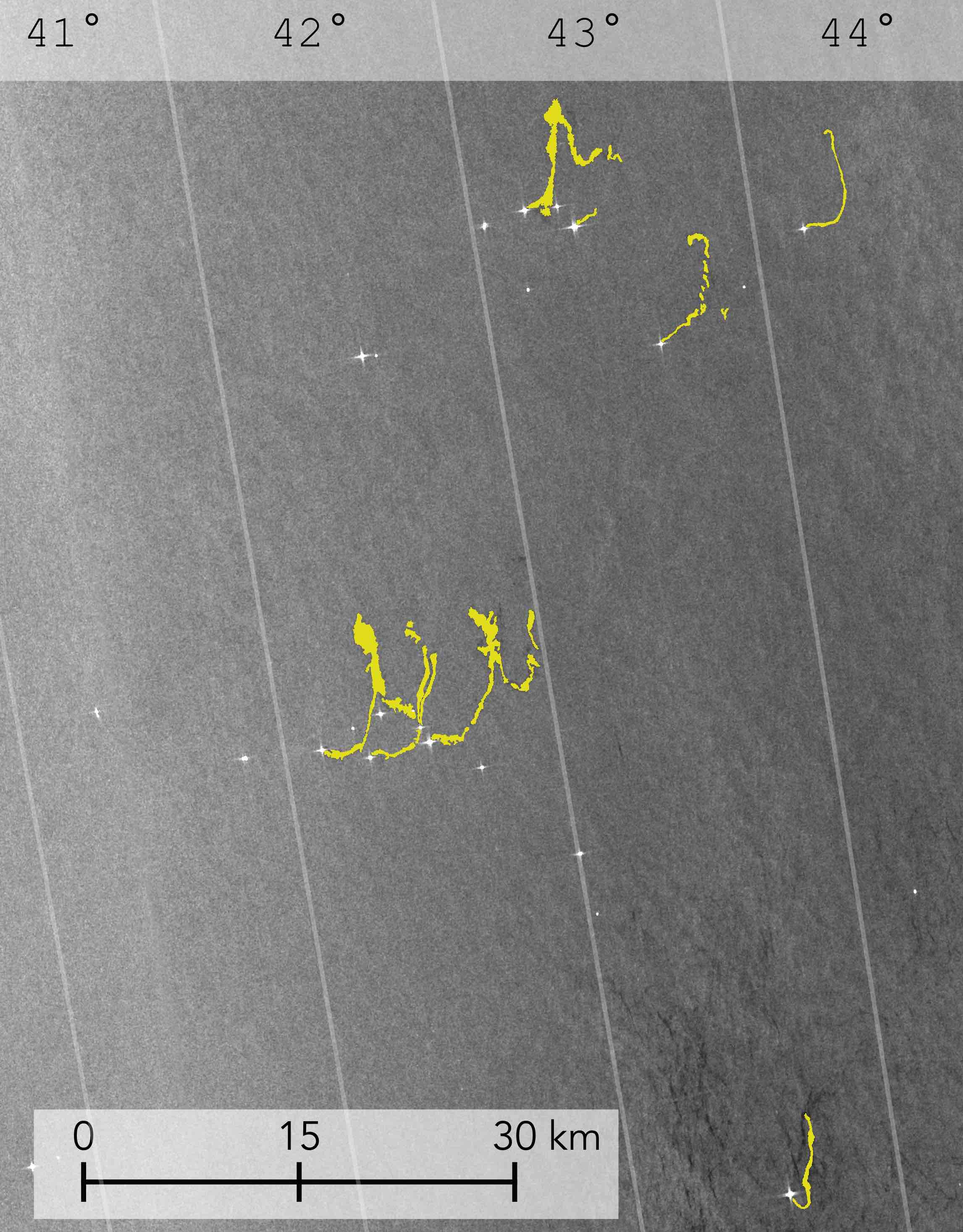}
    }\hspace{-0.7em}%
    ~
    \subfigure[Predicted mask]{
        \includegraphics[width=0.3\textwidth]{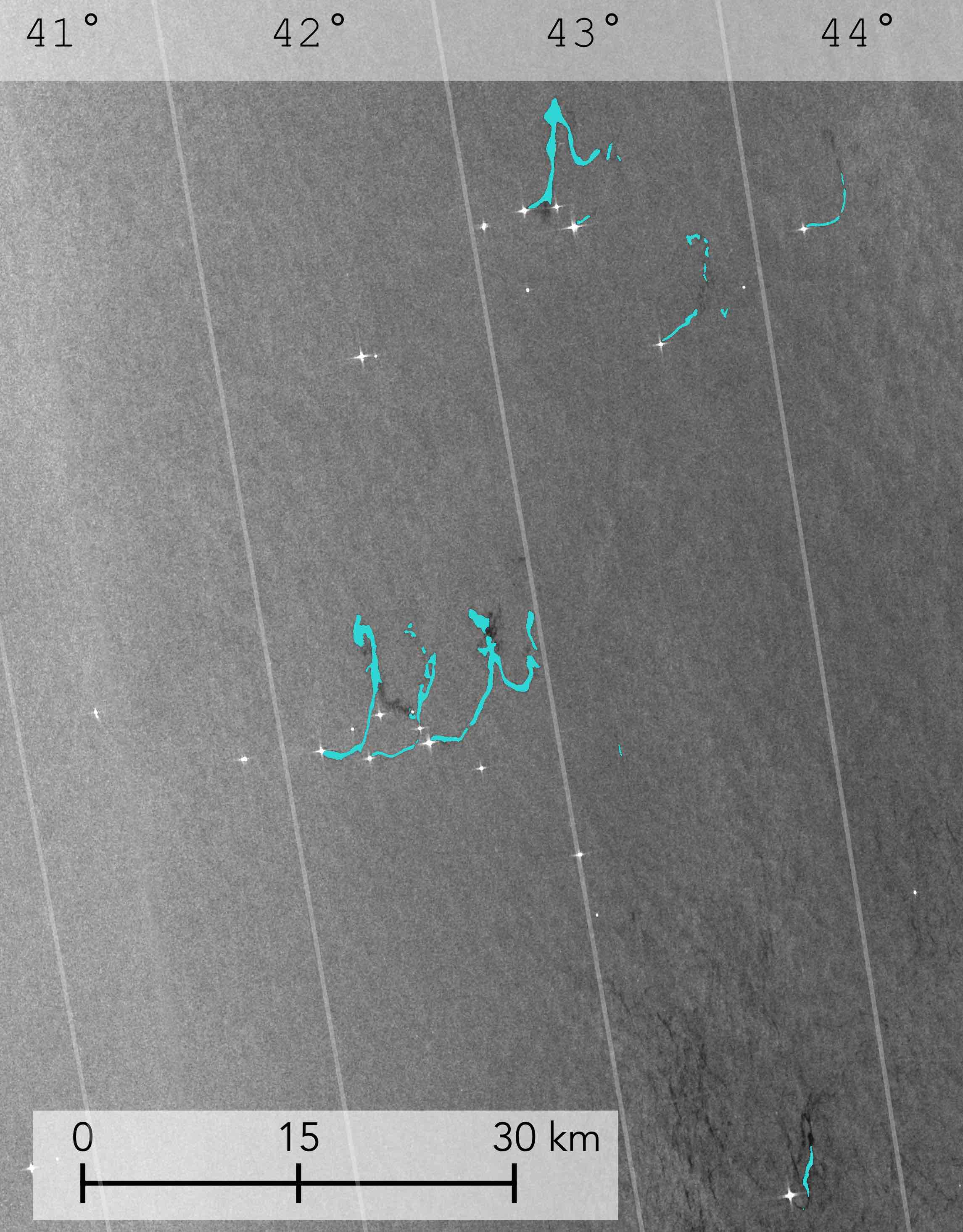}
    }\hspace{-0.7em}%

    \caption{Results on the 3 Sentinel-1 products used for testing. 
    Only small sections of the whole SAR products are shown in the figures. 
    The number on the top of the images represent the incident angle. From the left: original SAR image (VV-intensity), segmentation mask produced by a human operator (yellow masks), and segmentation mask predicted by the OFCN (blue masks).}
    \label{fig:seg_res}
\end{figure}

As discussed at the beginning of this section, OFCN returns a soft output in [0,1] which must be rounded to obtain a binary segmentation mask.
By varying the rounding threshold $\tau$ it is possible to vary quite significantly the number of TP, FP, and FN and also the two performance metrics, F1 score and IoU.
In particular, with a lower $\tau$ more FP appear, while a higher $\tau$ implies more FN. 
From Fig.~\ref{fig:test_thresh} we observe that in the three test products by using higher $\tau$ the number of FP decreases significantly and the FN only increases in the third image (Fig.~\ref{fig:test_thresh}(c)).
This suggests that a high $\tau$ value improves the detection precision.
On the other hand, the IoU and F1 score become much worse for high $\tau$ values, indicating that more precise contours can be found using a lower threshold.
We will exploit the behaviours observed for different $\tau$ when implementing our visualization tool, presented in Sec.~\ref{sec:nlive}.

Fig.~\ref{fig:seg_res} depicts examples of segmentation results on each one of the SAR products in $\mathcal{D}_\text{test}$.
Note the presence of a large FP in the second product (Fig.~\ref{fig:seg_res}(f)).

\subsection{Oil spill classification: experimental setting and results}

To train the classification network we first generate the predicted masks for both the training and validation set of $\mathcal{D}_1$ using the trained OFCN(32).
We decided to use the soft predictions, \textit{i.e.}, we do not threshold the output of the OFCN since it introduces an unnecessary bias and can conceal potential information of interest.
As previously discussed, the soft OFCN output values in [0,1] that can be interpreted as the \textit{amount of certainty} that a pixel belongs to the oil class.
Such a classification probability is lower in areas close to the edges of the oil spill, or where the oil starts to dissolve.

We trained 12 different instances of the classification models described in Sec.~\ref{sec:classification}, one for each category.
Each model is trained for $1,000$ epochs using Adam optimizer with initial learning rate $10^{-4}$, batch size $32$, L\textsubscript{2} norm regularization weight $10^{-6}$, and dropout $0.1$.
Image augmentation is used with the following parameters: max rotation 90$^{\circ}$, max-width shift 0.1 of total width, max height shift 0.1 of total height, max shearing 0.3, max zoom 0.2, probability of horizontal and vertical flips 0.5, pad mode ``mirror''.

The accuracies obtained for each category are reported in Tab.~\ref{tab:class_res}.
Since the values in most categories are unbalanced, we also report the F1 score as a performance measure.

\bgroup
\def\arraystretch{1} 
\begin{table}
\caption{Classification accuracy and F1 score for each one of the 12 categories on the validation set. 
}
\centering
\begin{tabular}{lcc}
\cmidrule[1.5pt]{1-3}
\textbf{Category} & \textbf{Accuracy} & \textbf{F1} \\ 
\cmidrule[.5pt]{1-3}
Patch shape         & 80.0 \% & 0.80  \\ 
Linear shape        & 76.8 \% & 0.77  \\ 
Angular shape       & 93.2 \% & 0.91  \\ 
Weathered texture   & 70.4 \% & 0.64  \\ 
Tailed texture      & 78.4 \% & 0.73  \\ 
Droplets texture    & 98.8 \% & 0.98  \\ 
Winding texture     & 94.4 \% & 0.92  \\ 
Feathered texture   & 97.2 \% & 0.96  \\ 
Shape outline       & 93.8 \% & 0.91  \\ 
Texture             & 55.6 \% & 0.49  \\ 
Contrast            & 61.6 \% & 0.59  \\ 
Edge                & 61.6 \% & 0.58  \\ 
\cmidrule[1.5pt]{1-3}
\end{tabular}
\label{tab:class_res}
\end{table}
\egroup

Compared to traditional image processing tools, CNNs usually achieve very high performance on recognizing textures and they exploit this capability to achieve high accuracy in downstream classification tasks. 
However, in our case, the performance obtained for some texture categories are particularly low.
We argue that one of the reasons is the presence of noise in the labelling process since it is difficult to precisely determine texture and contrast features from a SAR image in a consistent manner. 
We also observe a low accuracy and F1 score in some other categories, such as ``Contrast'' and ``Edge''.
Compared to the masks in the segmentation task, the classification labels are less reliable since the labelling procedure is more subjective and there is room for human errors. 
Most importantly, the trained operators use complementary information to define the categories and also to label them like an oil spill, such as sea state, wind, and historical seep sites.
The operators also account for nearby potential polluters (ships/platforms), by combining the automatic identification system (AIS) and sea maps.
Since all these information are not contained in the SAR products, a classification model based only on the image content can struggle in determining the right category and also to detect oil slicks that are not necessarily human-made.

\section{Large-scale visualization}
\label{sec:nlive}

We developed a pipeline to automatically acquire all the SAR products available in a given area and within a specified time frame and then process them with our deep learning framework.
Our pipeline performs the following steps:
\begin{enumerate}
    \item as input, we only specify the coordinates of an area and the time interval;
    \item from the Alaskan Sar Facility (ASF) repository all the SAR products within the time frame that overlaps at least 20\% with the specified area are fetched;
    \item since the SAR images come from Sentinel-1 (GDRH, 10m resolution), they are first smoothed and then down-sampled by a factor of 4 to match the mode of our training data;
    \item all the SAR products are processed with the OFCN described in Sec.~\ref{sec:detection}; the procedure consists of two steps, \textit{filtering} and \textit{coloring}, discussed below;
    \item each oil spill is associated with a vector of features, including the size of the slick and the distance from the closest oil spill detected;
    \item very small slicks are removed, \textit{i.e.}, slicks whose surface is lower than $0.25km^2$ and are further than $1.5km$ from any other oil spill.
\end{enumerate}

\begin{figure}[!ht]
    \centering
    \includegraphics[width=\textwidth]{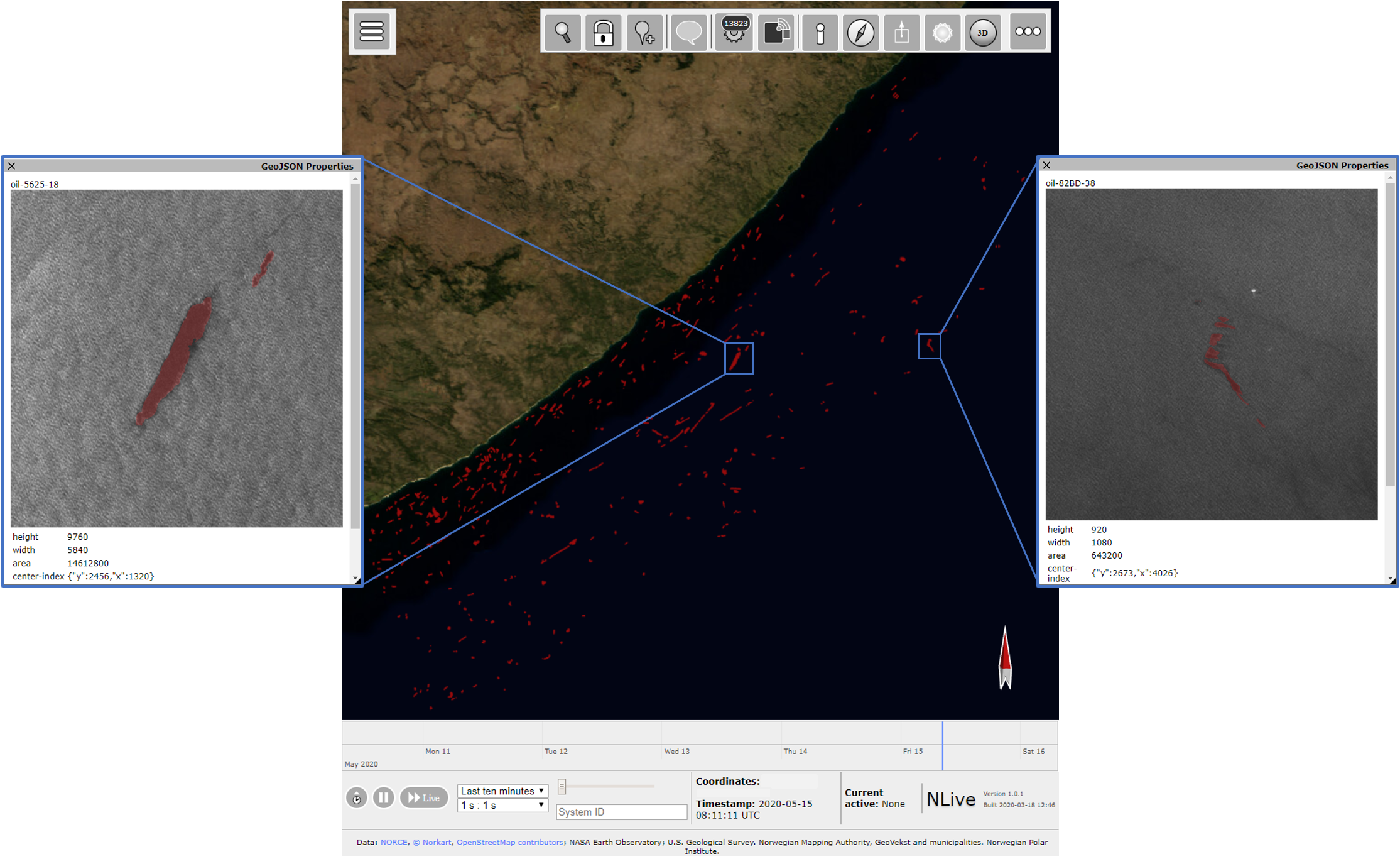}
    \caption{Visualization in NLive of oil spills detected in a large area (approximately $500 \times 200 km^2$) in the South hemisphere between 2014 and 2020.}
    \label{fig:nlive}
\end{figure}

Based on the discussion related to Fig.~\ref{fig:test_thresh}, it is possible to obtain high precision in the detection by thresholding the OFCN output with a high $\tau$. On the other hand, with a smaller $\tau$ the oil spill contours are more accurate.
For this reason, we first perform a \textit{filtering} applying $\tau=0.8$ to keep only larger slicks and discard many FPs.
In the \textit{coloring} step, we compute with $\tau=0.5$ the outline of the slicks that remain after filtering.
The \textit{filtering}-\textit{coloring} procedure is very fast since the soft output of OFCN does not need to be recomputed.

The results obtained are encoded into geojson files they are visualized with NLive\footnote{\url{http://nlive.norut.no}}, our geographic visualization tool.
In NLive it is possible to select the individual oil slicks and visualize a small chunk of SAR image where the oil spill has been detected, plus additional information about shape and the distance from the closest neighbour. 
The oil spills detected in an area of approximately $100,000\; km^2$ in the South hemisphere between October 2014 to March 2020 are shown in Fig.~\ref{fig:nlive}.
In the example, a total of 501 SAR products were retrieved, and in 136 of them at least one oil spill has been detected; a total of 665 oil spills were found.

\section{Conclusions}
\label{sec:conclusions}

In this paper, we proposed a deep learning framework to perform detection and categorization of oil spills on a large scale dataset.
We formulated oil spill detection as an image segmentation task, where each pixel in an input SAR image is assigned to the class ``oil'' or ``non-oil''. We designed a fully convolutional neural network for semantic segmentation, which we trained on pairs consisting of a small patch of a large SAR product and an associated binary mask, drawn by a human operator, that defines the class for each pixel. 
Through an extensive experimental evaluation, we demonstrated the capability of the proposed architecture in achieving high detection performance, obtaining results comparable to human operators. 

Once the oil spill is detected, we used a second neural network to classify an oil spill, according to 12 different categories describing shape and texture features. 
Differently from the detection task, the classification is not done at a pixel level but is relative to the whole patch. 
Our is the first exploratory work in categorizing oil spills in SAR images; the categorization results are useful to end-users and analysts to derive information about the source, stage of weathering and internal variations within the oil slicks that could be related to oil concentration or thickness. 

Despite neural networks are particularly capable of detecting textures, we obtained a low classification accuracy for some category.
We believe that part of the reason is the noise and inconsistency in the human-made labels. 
Indeed, it is extremely difficult to precisely determine texture and contrast features from a SAR image in a consistent manner. 
Remarkably, our findings on the automatic categorization performance provided valuable insights for improving the design of future oil spill services by operators such as KSAT.

Finally, we presented a production pipeline to detect and visualize the presence of oil spills worldwide at given times in history. 
Our pipeline fetches SAR products from the ASF repository of Sentinel-1 images and performs automatic detection and categorization.
The results are visualized in an interactive geographical map, where each oil spill can be individually selected to be further analyzed.
To the best of our knowledge, this is the first tool based on deep learning that allows analyzing oil activity on such a large scale.

\paragraph{Acknowledgements}
The Sentinel-1 source data is Copernicus Sentinel data, retrieved from ASF DAAC, processed by ESA. The work done by M. Espeseth is funded by CIRFA through the RCN (research grant no. 237906).
The work done by F. M. Bianchi and N. Borch is funded by KSAT, which is leading the Gonzales project 282082 of the PETROMAKS 2 program of the Norwegian research council (NFR).
We acknowledge the work of the following researchers at NORCE: Ingar Artnzen for processing and preparing the SAR dataset, Per Egil Kummervold for contributing to the design of deep learning methods, and Daniel Stødle as the main developer of NLive.
We also thank KSAT for providing us with the extensive dataset and for the discussions about the design of this study.

\bibliographystyle{abbrv}
{\footnotesize\bibliography{references}}

\appendix
\section{Further details of the segmentation model}
\label{sec:model_details}

\subsection{Bilinear upsampling layer}
A standard 2D upsampling procedure enlarges the image simply by inserting new rows and columns between the existing ones and fills them by replicating the content of the existing pixel values.
Instead, to obtain a more accurate generation of the output map we perform bilinear upsampling in the decoder layers of the OFCN.
Bilinear upsampling computes the new pixel values by performing a linear interpolation between the adjacent existing pixels. 
It has been shown that bilinear upsampling yields a more accurate reconstruction and the architecture equipped with it obtain a better segmentation accuracy~\cite{lin2016efficient}.

\subsection{Batch normalization layer}
To speed up the convergence of the training and provide a regularization to the network that improves its generalization capabilities, we applied Batch Normalization (BN)~\cite{ioffe2015batch} before each non-linear activation in the OFCN, both in the encoder and decoder.
BN normalizes channel-wise the mean and scale of the activations in the previous layer, making the network output almost invariant to the scale of the activations of the previous layers.
BN has the effect of reducing the variance in the distribution of layer activations over the course of training, preventing the network weights to diverge and the activation to saturate.
Empirically it was shown that BN stabilizes and accelerates the training while reducing the need to tune a variety of other hyperparameters to achieve higher performance~\cite{santurkar2018does}.

\subsection{Squeeze-and-Excitation layer}
We equipped the encoder of the OFCN with SE modules, which improve channel interdependencies at almost no additional computational cost~\cite{hu2018squeeze}.
SE gets a global understanding of each channel by squeezing each feature map to a single numerical value. 

\begin{figure}[!ht]
    \centering
    \includegraphics[keepaspectratio,width=0.6\textwidth]{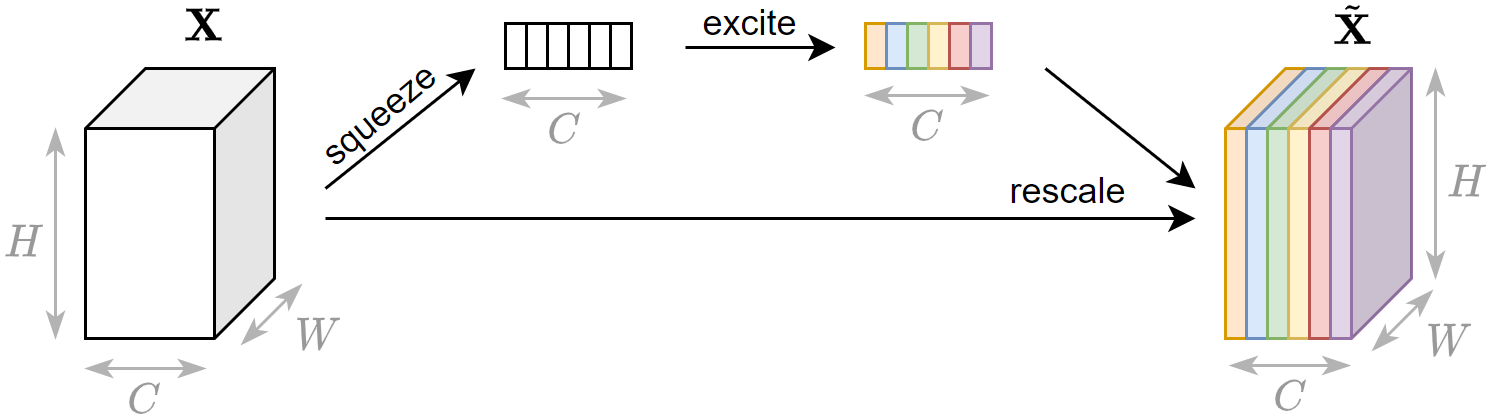}    
    \caption{Overview of the Squeeze-and-Excitation block used in the OFCN encoder. The SE blocks are inserted after each ReLU activation.}
    \label{fig:se}
\end{figure}

Fig.~\ref{fig:se} illustrates the mechanism of the Squeeze-and-Excitation block.
Let $\mathbf{X} \in \mathbb{R}^{H \times W \times C}$ be a feature map generated, for example, by a convolutional layer, where $H$ and $W$ are the height and width of the features maps and $C$ the number of channels (\textit{i.e.}, the neurons in the previous convolutional layer).
A squeeze operation aggregates the feature map $\mathbf{X}$ across the spatial dimensions $H$ and $W$.
The resulting embedding is a vector in $\mathbb{R}^{C}$ that captures the global distribution of channel-wise feature responses.
An excitation operation uses the embedding vector to implement a self-gating mechanism that rescales the weights of the original feature map channel-wise.
The resulting feature map $\tilde{\mathbf{X}}$ is used as input for the next neural network layer.

\subsection{Receptive field of the OFCN architecture}
\begin{figure}[!ht]
    \centering
    \includegraphics[keepaspectratio,width=.7\textwidth]{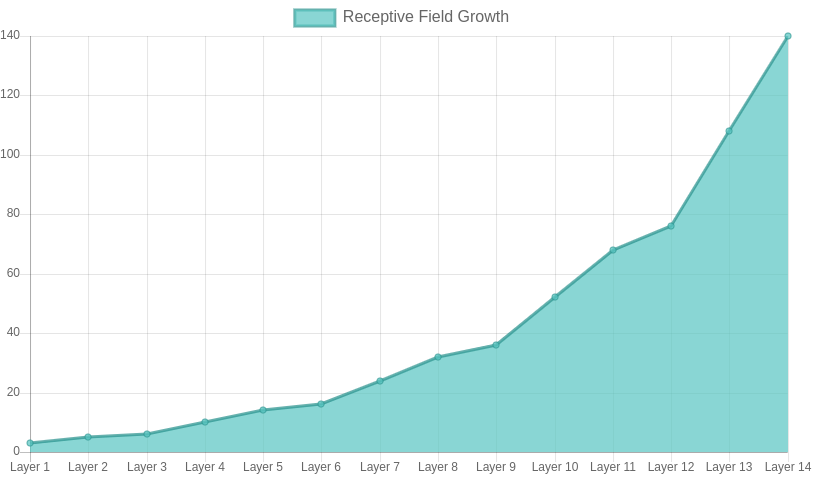}    
    \caption{Growth of the receptive field in the layers of the encoder.}
    \label{fig:receptive}
\end{figure}

Fig.~\ref{fig:receptive} depicts the exponential growth of the receptive field across the layers of the encoder in the proposed OFCN architecture to perform segmentation.
The diagram considers only the convolutional and pooling layers because are the only ones responsible for changing the size of the receptive field.

\section{Additional experimental details}
\label{sec:additional_details}

\subsection{Training stats of deep learning architectures} 

\begin{figure}[!h]
    \centering
    \subfigure[Training history of U-net]{
        \includegraphics[width=0.31\textwidth]{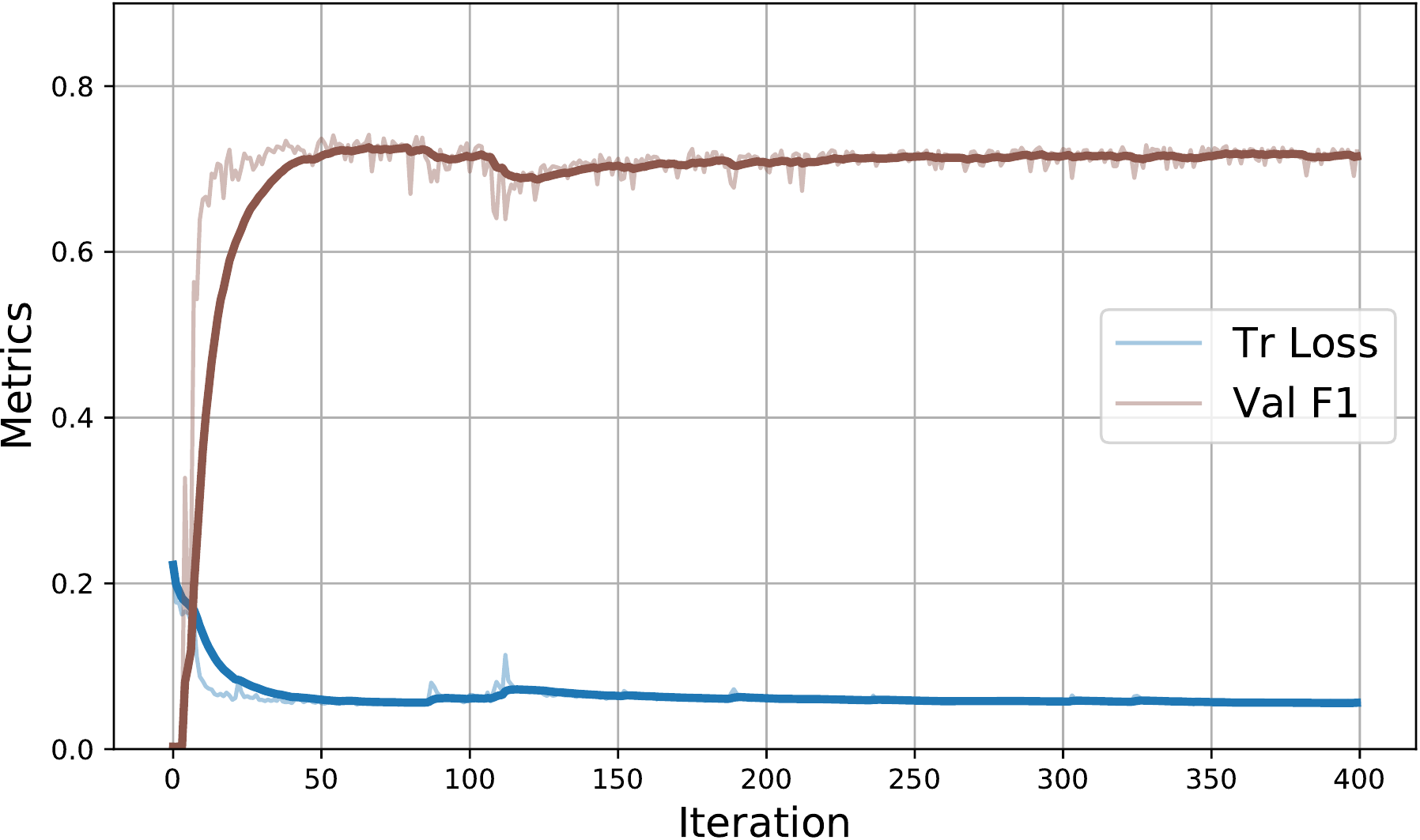}
    }\hspace{-0.5em}%
    ~
    \subfigure[Training history of DeeplabV3+]{
        \includegraphics[width=0.31\textwidth]{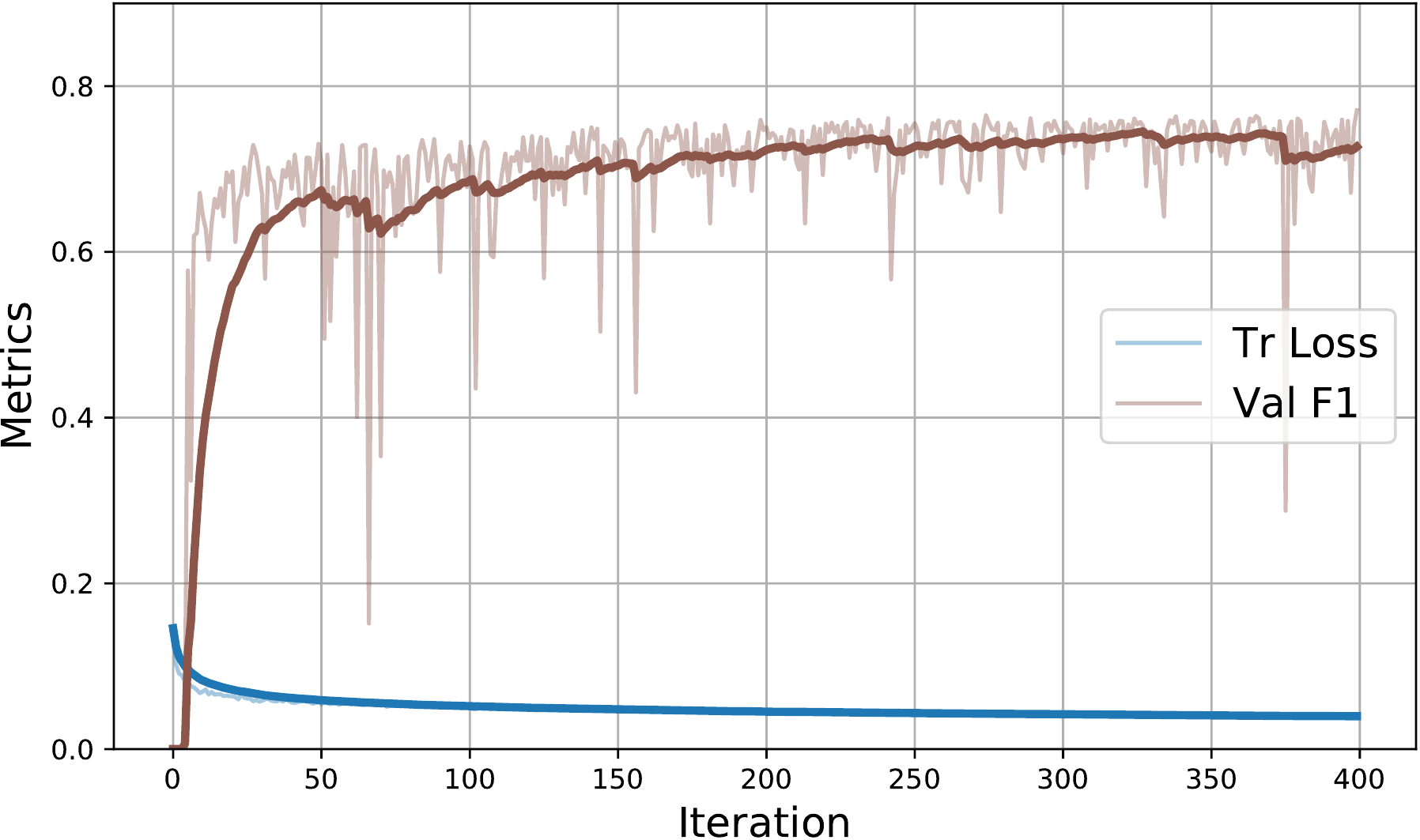}
    }\hspace{-0.5em}%
    ~
    \subfigure[Training history of OFCN]{
        \includegraphics[width=0.31\textwidth]{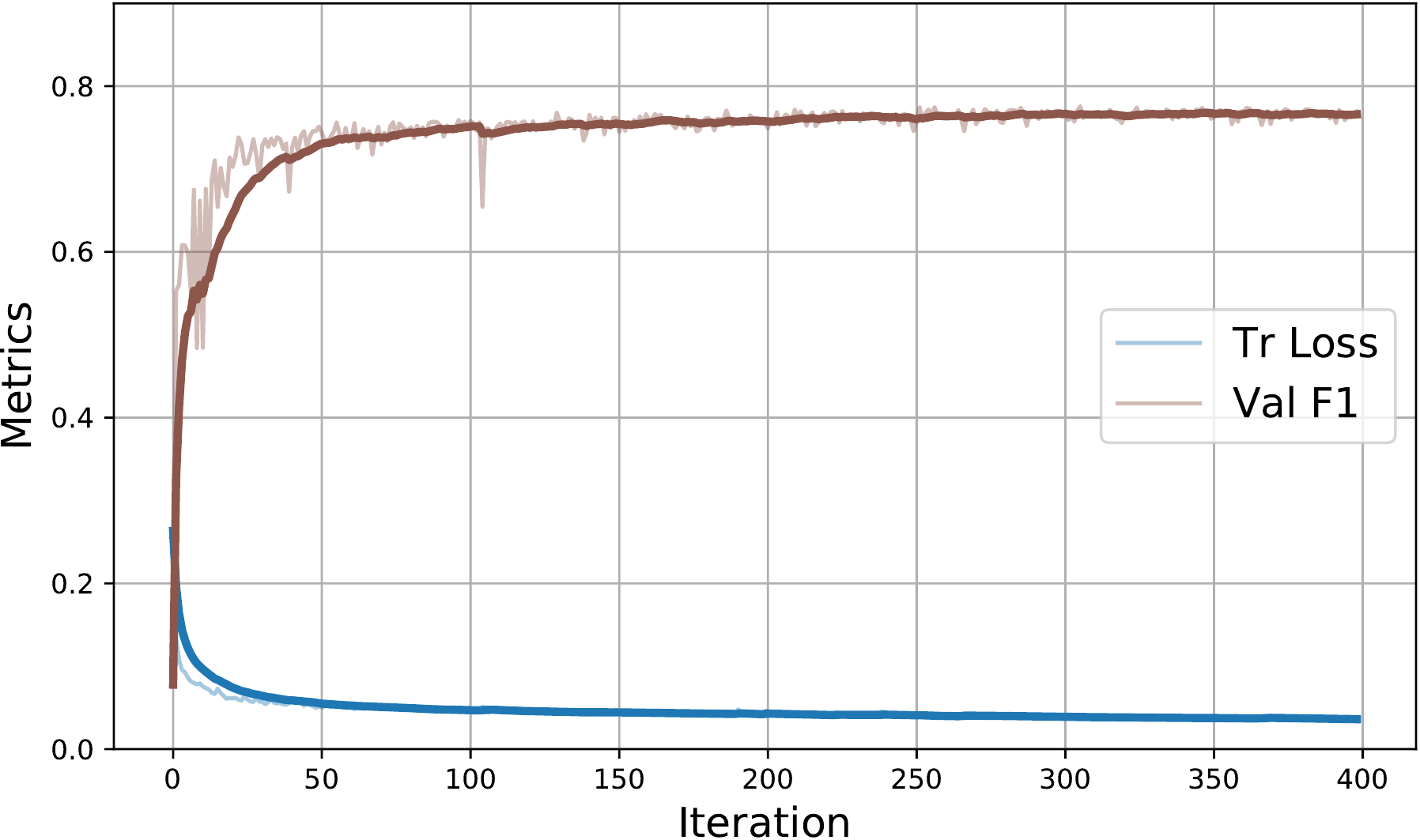}
    }\hspace{-0.5em}%
    \caption{Evolution of training loss and F1 score on validation across the 400 training epochs on dataset $\mathcal{D}_1$. 
    Bold line indicates a running average with window of size 30.}
    \label{fig:comparison_loss_plot}
\end{figure}

Fig~\ref{fig:comparison_loss_plot} depicts the evolution of the loss on the training set and the F1 score on the validation set during training, for the three deep learning models (U-net, DeeplabV3+, and the proposed OFCN) compared in Sec.~\ref{sec:detect_results}.
The models are trained on the dataset $\mathcal{D}_1$.
The plots show that none of the models is overfitting on the training set after 400 epochs.

\section{Unsuccessful approaches}
\label{sec:not_working}

In the following, we mention other strategies we experimented with but did not provide satisfactory results.

\subsection{Loss functions for class imbalance}

Besides the binary cross-entropy with class balancing, we tried three additional loss functions that are specifically designed to handle classes with an uneven number of samples.

\begin{itemize}
    \item \textit{Focal loss.} Addresses class imbalance by reshaping the standard cross-entropy loss such that it down-weights the loss assigned to well-classified examples~\cite{lin2017focal}. The Focal Loss focuses on training on a sparse set of hard examples and prevents the vast number of easy negatives from overwhelming the detector during training. 
    The Focal Loss is defined as
    \begin{equation}
        \label{eq:fc}
        FOC(p_t) = -\alpha(1-p_t)^{\gamma} \text{log}(p_t).
    \end{equation}
    We used the default parameters $\alpha = 0.25$ and $\gamma = 2$ proposed in the original paper.
    
    \item \textit{Jaccard Loss} handles class imbalance by computing the similarity between the predicted region and the ground-truth region for an object present in the image. In particular, the loss penalizes a naive algorithm that predicts every pixel of an image as the background, as the intersection between the predicted and ground-truth regions would be zero~\cite{rahman2016optimizing}.
    The Jaccard loss is defined as
    \begin{equation}
        \label{eq:jac}
         JAC(X,Y) = \frac{|X \cap Y|}{|X|+ |Y| - |X \cap Y|}
            =  \frac{\sum |X \odot Y|}{\sum |X| + \sum |Y| - \sum |X \odot Y|},
    \end{equation}
    where $\odot$ indicates the Hadamard product.

    \item \textit{Lov{\'a}sz-softmax loss} is an extension of the Jaccard Loss, which generates convex surrogates to submodular loss functions, including the Lovasz hinge. We refer to the original paper for the formal definition~\cite{berman2018lovasz}.
    The official TensorFlow implementation\footnote{\url{https://github.com/bermanmaxim/LovaszSoftmax}} has been used to perform the experiments.
    
\end{itemize}

For each loss function, we repeated the same hyperparameters search described in Sec.~\ref{sec:detect_results} and in Tab.~\ref{tab:losses} we report the best configuration found and the associated F1 score. 
%
\bgroup
\def\arraystretch{1} 
\setlength\tabcolsep{1em} 
\begin{table}[!ht]
\small
\centering
\begin{tabular}{lcccccc}
\cmidrule[1.5pt]{1-7}
\textbf{Loss} & \textbf{BN} & \textbf{SE} & \textbf{L\textsubscript{2} reg.} & \textbf{Dropout} & \textbf{LR} & \textbf{F1 ($\mathcal{D}_1$)} \\
\cmidrule[.5pt]{1-7}
JAC & True & True &  1e-6 & 0.1 & 1e-3 & 0.667 \\
FOC & True & True &  1e-3 & 0.0 & 1e-2 & 0.664 \\
LOV & True & True &  1e-5 & 0.0 & 1e-4 & 0.597 \\
\cmidrule[1.5pt]{1-7}
\end{tabular}
\caption{Best configurations and F1 scores for loss functions different from binary cross-entropy. \textit{Acronyms:} BN (Batch Normalization), SE (Squeeze-and-Excitation), L\textsubscript{2} reg. (strength of the L\textsubscript{2} regularization on the network parameters), LR (Learning Rate), FOC (Focal loss), JAC (Jaccard loss), LOV (Lov{\'a}sz-softmax loss). }
\label{tab:losses}
\end{table}
\egroup
%
It is immediately possible to notice that the results are significantly lower than those reported in Tab.~\ref{tab:crossval} and obtained by using binary cross-entropy and class weights.
In particular, when optimized with the Lov{\'a}sz-softmax loss, our model achieves an F1 score 17\% lower.

\subsection{Conditional Random Field.} 
The outcome of the prediction on each tile is modified by using CRF~\cite{krahenbuhl2011efficient} as a subsequent post-processing step.
CRF produces a result that is given by the combination of the pixel-wise neural network prediction, the pixel value in the input image (SAR value in this case) and pixel position.
More formally, the network prediction $\psi(x_i)$ for pixel $i$ is combined with the following pairwise potential
\begin{equation}
    \label{eq:crf}
    w^{(1)} \text{exp} \left( -\frac{|p_i - p_j|^2}{\theta_\alpha^2} -\frac{|I_i - I_j|^2}{\theta_\beta^2} \right) +  w^{(2)} \text{exp} \left( -\frac{|p_i - p_j|^2}{\theta_\gamma^2} \right)
\end{equation}
where $j$ are the indices of the other pixels in the patch, $p_i$ indicates pixel position and $I_i$ the SAR value of pixel $i$.
The parameters configuration used for the training is $w^{(1)} = 5$, $w^{(2)} = 0.1$, $\theta_\alpha^2 = 2$, $\theta_\beta^2 = 2$, and $\theta_\gamma^2 = 1$.

We found CRF to be computationally intensive, very sensitive to several hyperparameters that are difficult to tune, and, most importantly did bring significant improvement in the segmentation performance.
Similar results were found also in other related work~\cite{krestenitis2019early}.

\subsection{Multi-head classification network}
In a first attempt, to perform the categorization task we designed an architecture that shares the first 5 convolutional blocks and has 12 different output heads, each one specialized in predicting one of the 12 categories.
Such a network is capable of predicting at the same time all the categories given the input VV and mask.
However, we achieved better individual accuracy by training 12 different networks independently with a single head, one for each category, such as the one depicted in Fig.~\ref{fig:oilclass}.

\subsection{Gradient descent optimizers}
As an alternative to Adam, the Nadam optimizer~\cite{ruder2016overview} minimizes the loss faster in the first epochs, but in the end, settled worse to minima than Adam.


\end{document}

%% file: plot_data.txt
\def\patchDATA{(1,0.482)(2,0.518)}
\def\linearDATA{(1,0.618)(2,0.382)}
\def\angularDATA{(1,0.904)(2,0.096)} 
\def\weatheredDATA{(1,0.713)(2,0.287)} 
\def\tailedDATA{(1,0.832)(2,0.168)} 
\def\dropletsDATA{(1,0.983)(2,0.017)} 
\def\windingDATA{(1,0.925)(2,0.075)} 
\def\featheredDATA{(1,0.977)(2,0.023)} 
\def\shapeDATA{(1,0.784)(2, 0.216)}
\def\textureDATA{(1, 0.292)(2, 0.143)(3, 0.051)(4,0.512)}
\def\contrastDATA{(1, 0.230)(2, 0.528)(3, 0.241)}
\def\edgeDATA{(1, 0.337)(2, 0.133)(3, 0.529)}

%% file: arxiv_main.bbl
\begin{thebibliography}{10}

\bibitem{ALPERS2017133}
W.~Alpers, B.~Holt, and K.~Zeng.
\newblock Oil spill detection by imaging radars: Challenges and pitfalls.
\newblock {\em Remote Sens. Environ.}, 201:133--147, November 2017.

\bibitem{alpers1989}
W.~Alpers and H.~H{\"u}hnerfuss.
\newblock The damping of ocean waves by surface films: A new look at an old
  problem.
\newblock {\em J. Geophys. Res.}, 94(C5):6251--6265, May 1989.

\bibitem{Bakke2011}
T.~Bakke, A.~M.~V. Green, and P.~E. Iversen.
\newblock {\em Offshore Environmental Effects Monitoring in Norway --
  Regulations, Results and Developments}, pages 481--491.
\newblock Springer New York, New York, NY, 2011.

\bibitem{bengio2012unsupervised}
Y.~Bengio, A.~C. Courville, and P.~Vincent.
\newblock Unsupervised feature learning and deep learning: A review and new
  perspectives.
\newblock {\em CoRR, abs/1206.5538}, 1:2012, 2012.

\bibitem{berman2018lovasz}
M.~Berman, A.~Rannen~Triki, and M.~B. Blaschko.
\newblock The lov{\'a}sz-softmax loss: a tractable surrogate for the
  optimization of the intersection-over-union measure in neural networks.
\newblock In {\em Proceedings of the IEEE Conference on Computer Vision and
  Pattern Recognition}, pages 4413--4421, 2018.

\bibitem{bianchi2019snow}
F.~M. Bianchi, J.~Grahn, M.~Eckerstorfer, E.~Malnes, and H.~Vickers.
\newblock Snow avalanche segmentation in sar images with fully convolutional
  neural networks.
\newblock {\em arXiv preprint arXiv:1910.05411}, 2019.

\bibitem{sentinel1}
M.~Bourbigot, H.~Johnsen, and R.~Piantanida.
\newblock Sentinel-1 product definition.
\newblock Technical Report S1-RS-MDA-52-7440, Issue 2/7, MPC-S1, March 2016.
\newblock Online; accessed 14-June-2020.

\bibitem{BREKKE20051}
C.~Brekke and A.~H. Solberg.
\newblock Oil spill detection by satellite remote sensing.
\newblock {\em Remote Sens. Environ.}, 95(1):1 -- 13, 2005.

\bibitem{CANTORNA2019105716}
D.~Cantorna, C.~Dafonte, A.~Iglesias, and B.~Arcay.
\newblock Oil spill segmentation in sar images using convolutional neural
  networks. a comparative analysis with clustering and logistic regression
  algorithms.
\newblock {\em Applied Soft Computing}, 84:105716, 2019.

\bibitem{chen2018encoder}
L.-C. Chen, Y.~Zhu, G.~Papandreou, F.~Schroff, and H.~Adam.
\newblock Encoder-decoder with atrous separable convolution for semantic image
  segmentation.
\newblock In {\em Proceedings of the European conference on computer vision
  (ECCV)}, pages 801--818, 2018.

\bibitem{chollet2017xception}
F.~Chollet.
\newblock Xception: Deep learning with depthwise separable convolutions.
\newblock In {\em Proceedings of the IEEE conference on computer vision and
  pattern recognition}, pages 1251--1258, 2017.

\bibitem{Frate2000}
F.~{Del Frate}, A.~{Petrocchi}, J.~{Lichtenegger}, and G.~{Calabresi}.
\newblock Neural networks for oil spill detection using ers-sar data.
\newblock {\em IEEE Trans. Geosci. Remote Sens.}, 38(5):2282--2287, Sep. 2000.

\bibitem{ding2016convolutional}
J.~Ding, B.~Chen, H.~Liu, and M.~Huang.
\newblock Convolutional neural network with data augmentation for sar target
  recognition.
\newblock {\em IEEE Geoscience and remote sensing letters}, 13(3):364--368,
  2016.

\bibitem{dumoulin2016guide}
V.~Dumoulin and F.~Visin.
\newblock A guide to convolution arithmetic for deep learning.
\newblock {\em arXiv preprint arXiv:1603.07285}, 2016.

\bibitem{Gade1998}
M.~Gade, W.~Alpers, H.~H{\"u}hnerfuss, H.~Masuko, and T.~Kobayashi.
\newblock {Imaging of biogenic and anthropogenic ocean surface films by the
  multifrequency/multipolarization SIR-C/X-SAR}.
\newblock {\em J. Geophys. Res.}, 103(C9):18851--18866, August 1998.

\bibitem{Pineda2013}
O.~{Garcia-Pineda}, I.~R. {MacDonald}, X.~{Li}, C.~R. {Jackson}, and W.~G.
  {Pichel}.
\newblock Oil spill mapping and measurement in the gulf of mexico with textural
  classifier neural network algorithm (tcnna).
\newblock {\em IEEE J. Sel. Topics Appl. Earth Observ. Remote Sens.},
  6(6):2517--2525, Dec 2013.

\bibitem{ghosh2018stacked}
A.~Ghosh, M.~Ehrlich, S.~Shah, L.~S. Davis, and R.~Chellappa.
\newblock Stacked u-nets for ground material segmentation in remote sensing
  imagery.
\newblock In {\em CVPR Workshops}, pages 257--261, 2018.

\bibitem{Girard2005}
F.~Girard-Ardhuin, G.~Mercier, F.~Collard, and R.~Garello.
\newblock Operational {O}il-{S}lick {C}haracterization by {SAR} {I}magery and
  {S}ynergistic {D}ata.
\newblock {\em IEEE J. Oceanic Eng.}, 30(3):487--495, Jul. 2005.

\bibitem{goodfellow2016deep}
I.~Goodfellow, Y.~Bengio, and A.~Courville.
\newblock {\em Deep learning}.
\newblock MIT press, 2016.

\bibitem{holt2004a}
B.~Holt.
\newblock Chapter 2. {SAR} imaging of the ocean surface.
\newblock {\em Synthetic Aperture Radar Marine User's Manual (NOAA/NESDIS),
  C.R. Jackson and J. R. Apel}, pages 25--80, Sep. 2004.

\bibitem{howard2017mobilenets}
A.~G. Howard, M.~Zhu, B.~Chen, D.~Kalenichenko, W.~Wang, T.~Weyand,
  M.~Andreetto, and H.~Adam.
\newblock Mobilenets: Efficient convolutional neural networks for mobile vision
  applications.
\newblock {\em arXiv preprint arXiv:1704.04861}, 2017.

\bibitem{hu2018squeeze}
J.~Hu, L.~Shen, and G.~Sun.
\newblock Squeeze-and-excitation networks.
\newblock In {\em Proceedings of the IEEE conference on computer vision and
  pattern recognition}, pages 7132--7141, 2018.

\bibitem{ioffe2015batch}
S.~Ioffe and C.~Szegedy.
\newblock Batch normalization: Accelerating deep network training by reducing
  internal covariate shift.
\newblock {\em arXiv preprint arXiv:1502.03167}, 2015.

\bibitem{kingma2014adam}
D.~P. Kingma and J.~Ba.
\newblock Adam: A method for stochastic optimization.
\newblock {\em arXiv preprint arXiv:1412.6980}, 2014.

\bibitem{krahenbuhl2011efficient}
P.~Kr{\"a}henb{\"u}hl and V.~Koltun.
\newblock Efficient inference in fully connected crfs with gaussian edge
  potentials.
\newblock In {\em Advances in neural information processing systems}, pages
  109--117, 2011.

\bibitem{Krestenitis2019a}
M.~Krestenitis, G.~Orfanidis, K.~Ioannidis, K.~Avgerinakis, S.~Vrochidis, and
  I.~Kompatsiaris.
\newblock Early identification of oil spills in satellite images using deep
  cnns.
\newblock In I.~Kompatsiaris, B.~Huet, V.~Mezaris, C.~Gurrin, W.-H. Cheng, and
  S.~Vrochidis, editors, {\em MultiMedia Modeling}, pages 424--435, Cham, 2019.
  Springer International Publishing.

\bibitem{krestenitis2019early}
M.~Krestenitis, G.~Orfanidis, K.~Ioannidis, K.~Avgerinakis, S.~Vrochidis, and
  I.~Kompatsiaris.
\newblock Early identification of oil spills in satellite images using deep
  cnns.
\newblock In {\em International Conference on Multimedia Modeling}, pages
  424--435. Springer, 2019.

\bibitem{Krestenitis2019b}
M.~Krestenitis, G.~Orfanidis, K.~Ioannidis, K.~Avgerinakis, S.~Vrochidis, and
  I.~Kompatsiaris.
\newblock Oil spill identification from satellite images using deep neural
  networks.
\newblock {\em Remote Sensing}, 11(15), 2019.

\bibitem{landau2004handbook}
S.~Landau.
\newblock {\em A handbook of statistical analyses using SPSS}.
\newblock CRC, 2004.

\bibitem{li2018deepunet}
R.~Li, W.~Liu, L.~Yang, S.~Sun, W.~Hu, F.~Zhang, and W.~Li.
\newblock Deepunet: A deep fully convolutional network for pixel-level sea-land
  segmentation.
\newblock {\em IEEE J. Sel. Topics Appl. Earth Observ. Remote Sens.},
  11(11):3954--3962, 2018.

\bibitem{lin2016efficient}
G.~Lin, C.~Shen, A.~Van Den~Hengel, and I.~Reid.
\newblock Efficient piecewise training of deep structured models for semantic
  segmentation.
\newblock In {\em Proceedings of the IEEE Conference on Computer Vision and
  Pattern Recognition}, pages 3194--3203, 2016.

\bibitem{lin2017focal}
T.-Y. Lin, P.~Goyal, R.~Girshick, K.~He, and P.~Doll{\'a}r.
\newblock Focal loss for dense object detection.
\newblock In {\em Proceedings of the IEEE international conference on computer
  vision}, pages 2980--2988, 2017.

\bibitem{Minchew2012b}
B.~Minchew, C.~E. Jones, and B.~Holt.
\newblock Polarimetric {A}nalysis of {B}ackscatter {F}rom the {D}eepwater
  {H}orizon {O}il {S}pill {U}sing {L}-{B}and {S}ynthetic {A}perture {R}adar.
\newblock {\em IEEE Trans. Geosci. Remote Sens.}, 50(10):3812--3830, 2012.

\bibitem{nguyen2016synthesizing}
A.~Nguyen, A.~Dosovitskiy, J.~Yosinski, T.~Brox, and J.~Clune.
\newblock Synthesizing the preferred inputs for neurons in neural networks via
  deep generator networks.
\newblock In {\em Advances in neural information processing systems}, pages
  3387--3395, 2016.

\bibitem{NIVA2019}
N.~I. of~Water Reserach~(NIVA).
\newblock Environmental effects of offshore produced water discharges evaluated
  for the barents sea.
\newblock {\em Norwegian Environmental Agency (last access April 2020)}, 2019.

\bibitem{rahman2016optimizing}
M.~A. Rahman and Y.~Wang.
\newblock Optimizing intersection-over-union in deep neural networks for image
  segmentation.
\newblock In {\em International symposium on visual computing}, pages 234--244.
  Springer, 2016.

\bibitem{ronneberger2015u}
O.~Ronneberger, P.~Fischer, and T.~Brox.
\newblock U-net: Convolutional networks for biomedical image segmentation.
\newblock In {\em MICCAI}, 2015.

\bibitem{ruder2016overview}
S.~Ruder.
\newblock An overview of gradient descent optimization algorithms.
\newblock {\em arXiv preprint arXiv:1609.04747}, 2016.

\bibitem{Salberg2019}
A.~Salberg and S.~O. Larsen.
\newblock Classification of ocean surface slicks in simulated
  hybrid-polarimetric {SAR} data.
\newblock {\em IEEE Trans. Geosci. Remote Sens.}, 56(12):7062--7073, Dec 2018.

\bibitem{santurkar2018does}
S.~Santurkar, D.~Tsipras, A.~Ilyas, and A.~Madry.
\newblock How does batch normalization help optimization?
\newblock In {\em Advances in Neural Information Processing Systems}, pages
  2483--2493, 2018.

\bibitem{Singh1986}
K.~P. Singh, A.~L. Gray, R.~K. Hawkins, and R.~A. O'Neil.
\newblock {The Influence of Surface Oil on C-and Ku-Band Ocean Backscatter}.
\newblock {\em IEEE Trans. Geosci. Remote Sens.}, GE-24(5):738 -- 744,
  September 1986.

\bibitem{Singha2013}
S.~{Singha}, T.~J. {Bellerby}, and O.~{Trieschmann}.
\newblock Satellite oil spill detection using artificial neural networks.
\newblock {\em IEEE J. Sel. Topics Appl. Earth Observ. Remote Sens.},
  6(6):2355--2363, Dec 2013.

\bibitem{Skrunes2014}
S.~Skrunes, C.~Brekke, and T.~Eltoft.
\newblock {Characterization of Marine Surface Slicks by Radarsat-2
  Multipolarization Features}.
\newblock {\em IEEE Trans. Geosci. Remote Sens.}, 52(9):5302--5319, September
  2014.

\bibitem{Skrunes2019}
S.~Skrunes, A.~M. Johansson, and C.~Brekke.
\newblock Synthetic aperture radar remote sensing of operational platform
  produced water releases.
\newblock {\em Remote Sensing}, 11(23), 2019.

\bibitem{Solberg1999}
A.~H.~S. {Solberg}, G.~{Storvik}, R.~{Solberg}, and E.~{Volden}.
\newblock Automatic detection of oil spills in ers sar images.
\newblock {\em IEEE Trans. Geosci. Remote Sens.}, 37(4):1916--1924, July 1999.

\bibitem{srivastava2014dropout}
N.~Srivastava, G.~Hinton, A.~Krizhevsky, I.~Sutskever, and R.~Salakhutdinov.
\newblock Dropout: a simple way to prevent neural networks from overfitting.
\newblock {\em The journal of machine learning research}, 15(1):1929--1958,
  2014.

\bibitem{TOPOUZELIS2007264}
K.~Topouzelis, V.~Karathanassi, P.~Pavlakis, and D.~Rokos.
\newblock Detection and discrimination between oil spills and look-alike
  phenomena through neural networks.
\newblock {\em ISPRS Journal of Photogrammetry and Remote Sensing}, 62(4):264
  -- 270, 2007.

\bibitem{zhao2017survey}
B.~Zhao, J.~Feng, X.~Wu, and S.~Yan.
\newblock A survey on deep learning-based fine-grained object classification
  and semantic segmentation.
\newblock {\em International Journal of Automation and Computing},
  14(2):119--135, 2017.

\bibitem{zheng2018feature}
A.~Zheng and A.~Casari.
\newblock {\em Feature engineering for machine learning: principles and
  techniques for data scientists}.
\newblock " O'Reilly Media, Inc.", 2018.

\bibitem{zhu2017deep}
X.~X. Zhu, D.~Tuia, L.~Mou, G.-S. Xia, L.~Zhang, F.~Xu, and F.~Fraundorfer.
\newblock Deep learning in remote sensing: a comprehensive review and list of
  resources.
\newblock {\em IEEE Geoscience and Remote Sensing Magazine}, 2017.

\end{thebibliography}
